\documentclass[sigconf]{acmart}
\AtBeginDocument{%
  \providecommand\BibTeX{{%
    \normalfont B\kern-0.5em{\scshape i\kern-0.25em b}\kern-0.8em\TeX}}}

\usepackage{soul}
\usepackage{graphicx}
\usepackage{booktabs}
\usepackage{float}
\usepackage{overpic}
\usepackage{float}  %设置图片浮动位置的宏包
\usepackage{subfloat}
\usepackage{stfloats} % for pic
\usepackage[skip=0.5\baselineskip]{caption}
\usepackage{bm}
\usepackage{amsmath}
\usepackage{booktabs}
\usepackage{multirow}
\usepackage{multicol}
\usepackage{enumitem}
\usepackage{subcaption}
\usepackage{threeparttable}
\usepackage{xspace}
\usepackage{color}
\usepackage[normalem]{ulem}
\usepackage{algorithmicx,algorithm}
\usepackage{algpseudocode}
\usepackage{algorithm,algpseudocode}
\usepackage{marvosym}   % 添加通讯作者的信封
\usepackage{bm}
\usepackage{makecell}

\makeatletter
\def\@ACM@checkaffil{% Only warnings
    \if@ACM@instpresent\else
    \ClassWarningNoLine{\@classname}{No institution present for an affiliation}%
    \fi
    \if@ACM@citypresent\else
    \ClassWarningNoLine{\@classname}{No city present for an affiliation}%
    \fi
    \if@ACM@countrypresent\else
        \ClassWarningNoLine{\@classname}{No country present for an affiliation}%
    \fi
}
\makeatother

\usepackage{xspace}
\newcommand{\etal}{\emph{et al.}\xspace}
\newcommand{\eg}{\emph{e.g.,}\xspace}
\newcommand{\ie}{\emph{i.e.,}\xspace}
\newcommand{\etc}{\emph{etc.}\xspace}

\copyrightyear{2024}
\acmYear{2024}
\setcopyright{acmlicensed}
\acmConference[SIGIR '24]{Proceedings of the 47th International ACM SIGIR Conference on Research and Development in Information Retrieval}{July 14--18, 2024}{Washington, DC, USA}
\acmBooktitle{Proceedings of the 47th International ACM SIGIR Conference on Research and Development in Information Retrieval (SIGIR '24), July 14--18, 2024, Washington, DC, USA}
\acmDOI{10.1145/3626772.3657722}
\acmISBN{979-8-4007-0431-4/24/07}

\emergencystretch=\maxdimen
\begin{document}
\begin{sloppypar}   % 自动换行和对齐，防止一行文字溢出
\title{When MOE Meets LLMs: Parameter Efficient Fine-tuning for Multi-task Medical Applications}

%%
%% The "author" command and its associated commands are used to define the authors and their affiliations.
%% Of note is the shared affiliation of the first two authors and the
%% "authornote" and "authornotemark" commands used to denote shared contribution to the research.
\author{Qidong Liu}
%\email{webmaster@marysville-ohio.com}
\affiliation{%
  \institution{Xi'an Jiaotong University \& \\ City University of Hong Kong}
  \city{Xi'an}
  % \state{Shaanxi}
  \country{China}
}
\email{liuqidong@stu.xjtu.edu.cn}

\author{Xian Wu \Letter}
\thanks{\Letter \ \text{Corresponding authors}}
\affiliation{%
  \institution{Jarvis Research Center, \\ Tencent YouTu Lab}
  \city{Shenzhen}
  % \state{Guangdong}
  \country{China}
}
\email{kevinxwu@tencent.com}

\author{Xiangyu Zhao \Letter}
\affiliation{%
  \institution{City University of Hong Kong}
  \city{Hong Kong}
  \country{Hong Kong}
}
\email{xianzhao@cityu.edu.hk}

\author{Yuanshao Zhu}
\affiliation{%
  \institution{Southern University of Science and Technology \& \\ City University of Hong Kong}
  \city{Shenzhen}
  % \state{Guangdong}
  \country{China}
}
\email{zhuys2019@mail.sustech.edu.cn}

\author{Derong Xu}
\affiliation{%
  \institution{University of Science and Technology of China \& \\ City University of Hong Kong}
  \city{Hefei}
  % \state{Anhui}
  \country{China}
}
\email{derongxu@mail.ustc.edu.cn}

\author{Feng Tian \Letter}
\affiliation{%
  \institution{Xi'an Jiaotong University}
  \city{Xi'an}
  % \state{Shaanxi}
  \country{China}
}
\email{fengtian@mail.xjtu.edu.cn}

\author{Yefeng Zheng}
\affiliation{%
  \institution{Jarvis Research Center, \\ Tencent YouTu Lab}
  \city{Shenzhen}
  % \state{Guangdong}
  \country{China}
}
\email{yefengzheng@tencent.com}

%%
%% By default, the full list of authors will be used in the page headers. Often, this list is too long, and will overlap other information printed in the page headers. This command allows the author to define a more concise list of authors' names for this purpose.
\renewcommand{\shortauthors}{Qidong Liu \etal}

%% The abstract
\begin{abstract}
% The recent surge in large language models (LLMs) has garnered significant interest across numerous fields, e.g., medical applications. Some efforts fine-tune LLMs to equip these language models with medical domain-specific knowledge. However, two challenges arise in this process. The first challenge is the high tuning cost, which results from the extensive number of parameters in LLMs. Although several parameter-efficient fine-tuning techniques have been proposed, they are often hindered by the second challenge, i.e., the scarcity of medical data.

  The recent surge in Large Language Models (LLMs) has garnered significant attention across numerous fields. Fine-tuning is often required to fit general LLMs for a specific domain, like the web-based healthcare system. 
  However, two problems arise during fine-tuning LLMs for medical applications. 
  One is the task variety problem, which involves distinct tasks in real-world medical scenarios. The variety often leads to sub-optimal fine-tuning for data imbalance and seesaw problems.
  %Fine-tuning LLMs for each task is feasible but obviously laborious and skillful. Though some unified multi-task learning frameworks have been proposed to alleviate such a problem, their benefits are shed by the other problem, \ie unbearable tuning cost. 
  Besides, the large amount of parameters in LLMs leads to huge time and computation consumption by fine-tuning.
  To address these two problems, we propose a novel parameter efficient fine-tuning framework for multi-task medical applications, dubbed as \textbf{MOELoRA}. 
  The designed framework aims to absorb both the benefits of mixture-of-expert (MOE) for multi-task learning and low-rank adaptation (LoRA) for parameter efficient fine-tuning. 
  For unifying MOE and LoRA, we devise multiple experts as the trainable parameters, where each expert consists of a pair of low-rank matrices to retain the small size of trainable parameters. 
  Then, a task-motivated gate function for all MOELoRA layers is proposed, which can control the contributions of each expert and produce distinct parameters for various tasks. 
  We conduct experiments on a multi-task medical dataset, indicating MOELoRA outperforms the existing parameter efficient fine-tuning methods.
  %, and can combine the task-shared and --specific knowledge.
  The code is available online~\footnote{\url{https://github.com/Applied-Machine-Learning-Lab/MOELoRA-peft}}.

\end{abstract}

%% CCS Concepts



%% This command processes the author affiliation and title
\maketitle

\section{Introduction}

%%% The background and importance of fine-tuning LLMs for Medical Applications %%%
%%% To illustrate fine-tuning for medical applications is necessary 
Due to the impressive capabilities in language understanding and generation, the Large Language Models (LLMs), such as ChatGPT~\cite{liu2023gpt} and ChatGLM~\cite{zeng2022glm}, have gained extensive interest from both academia and industry. Many efforts have been devoted to investigating the potential applications of LLMs across various domains~\cite{fan2023recommender,wang2023survey,hadi2023survey}. One particularly suitable domain for LLMs is the medical domain, as the application of LLMs can benefit both patients and doctors. For patients, the LLM-enabled online Chatbot can provide convenient access to medical knowledge; For doctors, the LLM-enabled Clinical Decision Supporting Systems (CDSS) can relieve heavy workloads and improve diagnosis efficiency.
% The online medical systems, given their critical importance, have naturally drawn significant interest in integration with LLMs.

However, the majority of LLMs are trained for general purposes and are not customized for the medical domain. As a result, the general LLMs often fall short in medical tasks due to a lack of specialized medical knowledge~\cite{yunxiang2023chatdoctor}. 
To empower LLMs with medical capabilities, a straightforward manner is to fine-tune LLMs with medical tasks.
For large LLMs with more than $100$ billion parameters, they are usually closed-source and extremely costly for fine-tuning~\cite{openai2023gpt4}. %\zys{cite}. T
% This poses challenges for their widespread adoption in clinical settings. 
Therefore, in this paper, we focus on the open-source LLMs and fine-tuning them with medical knowledge and clinical tasks~\cite{wang2023huatuo,yunxiang2023chatdoctor}. 
% Pioneering efforts in this direction have already been made, as evidenced by recent studies~\cite{wang2023huatuo,yunxiang2023chatdoctor}.

Fine-tuning LLMs for the medical domain usually involves two primary challenges:
(i) \textbf{Task Variety Problem}: In real-world clinics, LLMs can be applied to a large range of tasks, like doctor recommendation~\cite{doctor2022}, diagnosis prediction~\cite{Qiao2019MNNMA}, medicine recommendation~\cite{yingyingtois,zheng2023interaction}, medical named entity recognition~\cite{zhao2019neural,rezayi2022clinicalradiobert}, clinical report generation~\cite{miura2020improving} and \etc Since the input and output of these tasks are quite different, it is difficult to fine-tune a unified model for all tasks.
Given the diversity of these tasks, fine-tuning a single model for each specific task is feasible but demands extensive expertise and labor. 
An integrated multi-task learning framework could potentially address this issue. 
However, much of the existing research on LLMs, as seen in studies like~\cite{wang2023huatuo,singhal2022large}, predominantly centers on medical dialogue. 
Such over-attention ignores the variety of tasks, resulting in multi-task fine-tuning remains underexplored. %\lqd{cite some medical papers (nature)}
(ii) \textbf{High Tuning Cost}: 
While fine-tuning all model parameters was a standard approach during the era of Bert~\cite{kenton2019bert}, it becomes challenging for LLMs due to their sheer size. 
The vast number of parameters in LLMs can lead to prohibitive time and computational expenses in practice~\cite{yang2023baichuan}. %\zxy{is it severer in medical application?}
As such, there is an urgent need for parameter efficient fine-tuning methodologies.
To address these two challenges, the community urgently calls for developing a multi-task parameter efficient fine-tuning framework for LLM-driven medical applications.
% To address these two challenges， we explore a \textit{\textbf{multi-task parameter efficient fine-tuning framework}} for the LLM-driven medical applications.
% \zys{你这里说为了解决两项挑战，所以xxx, 但是后面就没了，反而去介绍多任务学习了？}
%\lqd{has much healthcare costs. LLM will make the burden server}

\begin{figure}[!t]
\centering
% \vspace{-1mm}
\includegraphics[width=0.9\linewidth]{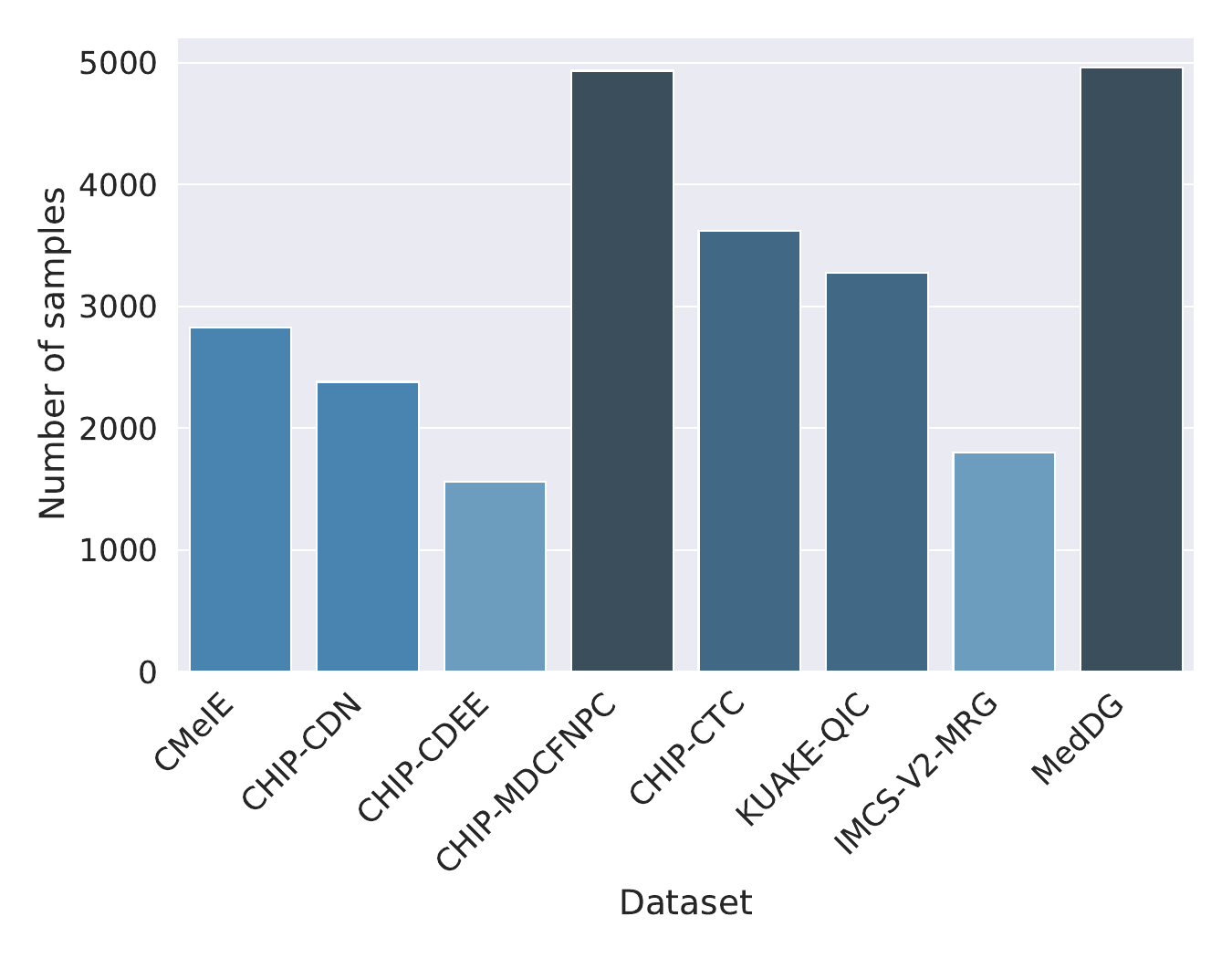}
\caption{The illustration for data imbalance problem.}
\label{fig:pre_task}
\vspace{-5mm}
\end{figure}

%%% The problem faced by fine-tuning LLMs for medical applications %%%
% illustrate we use the LLM to complete specific medical, but not only the dialogue
For the task variety problem, several multi-task learning frameworks have been proposed~\cite{zhang2021survey,crawshaw2020multi,wang2023multi,fan2022comprehensive,liu2023multi}. 
A standout among these is Mixture-of-Experts (MOE)~\cite{shazeer2017outrageously}, which designs multiple separate experts to learn task-shared and -specific knowledge, and integrates a gate function to modulate the contributions of each expert. 
While existing frameworks adeptly consolidate multiple tasks for classical neural network architectures, they are primarily compatible with full fine-tuning, which is associated with high tuning costs.
Correspondingly, the emergence of parameter efficient fine-tuning (PEFT) methods has offered a potential solution to the problem of high tuning costs. 
These methods typically tune a limited number of parameters, keeping the pre-trained LLMs parameters frozen. 
% For instance, LoRA~\cite{hu2021lora} proposes to only train pairs of low-rank matrices for fitting the parameter updates of dense layers in LLMs.
%However, a limitation of PEFT is its tendency to fine-tune a singular set of parameters across all tasks.
However, the existing PEFT is limited to fine-tuning either multiple sets of parameters for each task separately or a singular set across all tasks. 
Though separate training can fit each task well, this strategy is laborious and lacks task-shared knowledge. 
While fine-tuning a set of parameters is feasible, it may hurt performance due to issues such as data imbalance and seesaw effects~\cite{crawshaw2020multi,li2020dice}.
For illustration, we analyze the data distribution of a Chinese medical dataset, PromptCBLUE~\cite{zhu2023promptcblue}, in Figure~\ref{fig:pre_task}. 
Our analysis reveals significant disparities: while some tasks boast nearly $5,000$ samples, others have fewer than $2,000$. 
This imbalance can skew the uniquely fine-tuned parameters towards tasks with more samples, inadvertently undermining the performance on tasks with fewer samples. 
\textit{Therefore, parameter efficient fine-tuning of separate parameters for multi-task by a unique training process can alleviate both problems.}
%\lqd{Since neither of these two existing directions cannot handle the two problems simultaneously, we aim to propose the multi-task PEFT to target it.} 
% \zys{这个段落并没有起到承上启下的作用，有点像文献综述又像背景介绍}

%% 单栏图
%%% Task Distribution figure %%%
%%% Task Distribution figure %%%

%%% The technical challenge for the combination of LoRA and MOE %%%
%Considering the benefits of PEFT and multi-task learning, we aim to collaborate on their respective strengths, which are applied to address the challenges of task variety and high tuning costs.
To address the challenges of task variety and high tuning costs, we propose a unified parameter efficient fine-tuning framework to learn separate parameters for various tasks, dubbed as MOELoRA.
%LoRA has been demonstrated as an efficient method for fine-tuning LLMs~\cite{hu2021lora}, so our approach follows the major process of it. 
Our framework follows the basic scheme of LoRA for the parameter efficiency, \ie only fine-tuning small size of parameters parallel to the dense layers in LLMs.
However, as discussed previously, existing unified LoRA fine-tuning faces the challenge of \textbf{a singular set of parameters across all tasks}. Thus, in our approach, we first design several experts as the trainable part rather than a singular pair of low-rank matrices. On the one hand, inspired by MOE~\cite{shazeer2017outrageously}, separate experts can help learn task-specific knowledge under one unique training process. On the other hand, such a design gives the chance to produce several distinct sets of parameters.
Besides, for parameter efficiency, we devise each expert as two low-rank matrices. Then, to learn separate sets of parameters for each task, we propose a task-motivated gate function. In specific, the gate function absorbs the task identity and outputs corresponding expert weights. By the expert weights for one specific task and the parameters of multiple experts, we can get the unique updated parameters for this task.
%, but it remains susceptible to the data imbalance and seesaw issues previously discussed. 
% Our approach seeks to integrate the MOE concept into LoRA. However, this amalgamation is a non-trivial task.
% Firstly, the multiple experts are the only way in the forward process, which is different from the two-thread forward process of LoRA, \ie pre-trained parameters and low-rank matrices.
% To harmonize the disparate forward processes of LoRA and MOE, we reformulate the forward process in LoRA as a pre-trained parameter and MOE, where the MOE contains several low-rank matrix pairs as the expert.
% Given that the expert structure remains a low-rank matrix, we can ascertain that there is no additional increase in the size of trainable parameters. 
% We refer to this novel architecture as \textbf{MOELoRA} throughout the paper.
% The second challenge stems from the utilization of \textbf{a singular set of parameters across all tasks}, complicating the acquisition of task-specific information. 
% To address this, we introduce a task-motivated gate function, which can learn task-specific expert weights to produce distinct updated parameters for each task.
% Similar to LoRA, we can retrieve the updated parameters for individual tasks, ensuring that no additional inference latency is introduced when deploying the LLM for specific tasks.

% The expert weights vary according to the input sample in the original MOE, which causes the additional inference latency

%%% Contributions %%%
In summary, the contributions of this paper are as follows:
\begin{itemize}[leftmargin=*]

    \item We introduce \textbf{MOELoRA}, a novel multi-task PEFT framework that combines the strengths of both MOE and LoRA. Additionally, we design a task-motivated gate function to facilitate the tuning of distinct parameter sets for each task.

    \item We conduct comprehensive experiments on a public multi-task Chinese medical dataset, with the results underscoring the superiority of the proposed MOELoRA framework.

    \item To our knowledge, this research represents the first endeavor to delve into multi-task parameter efficient fine-tuning techniques for LLM-driven medical applications.
    
\end{itemize}

\section{Preliminary}

%% 单栏图
%%% LLMs for Medical Applications %%%
\begin{figure}[!t]
\centering
% \vspace{-1mm}
\includegraphics[width=1\linewidth]{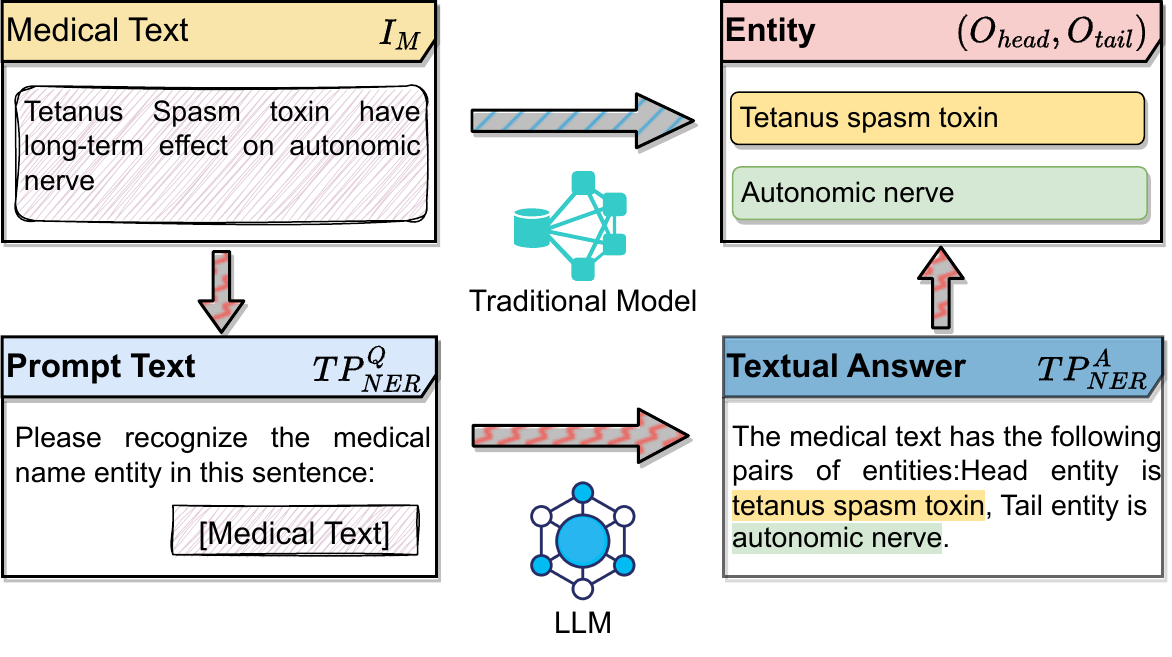}
\caption{The medical name entity recognition example for illustration of how to use LLMs to complete medical tasks.}
\label{fig:pre_llm}
\vspace{-6mm}
\end{figure}
%%% LLMs for Medical Applications %%%

% In this section, we first briefly introduce how LLMs are adopted for medical applications.
% Then, we give the problem definition of multi-task medical applications.

\subsection{LLMs for Medical Applications} \label{sec:preliminary_llm}
Intelligent medical systems have become increasingly prevalent in contemporary web-based healthcare settings. 
Numerous studies have sought to standardize medical tasks by defining consistent input and output patterns, thereby streamlining the model design process. 
As the example of medical named entity recognition (NER)~\cite{zhao2019neural,rezayi2022clinicalradiobert} illustrated in Figure~\ref{fig:pre_llm}, traditional models typically process medical texts, denoted as $I_M$, to produce entities $O_{head}$ and $ O_{tail}$.
However, the integration of LLMs into medical tasks introduces a distinct paradigm. 
Given that both the input and output of LLMs are typically linguistic in nature, there is a necessity to reformulate medical tasks to be compatible with LLMs.

To adapt medical tasks for LLMs, we need to restructure both the input and output patterns.
\textbf{Input Modification}: We incorporate instruction templates into the original medical texts to guide LLMs in executing the relevant tasks~\cite{zhang2023instruction}. 
Taking medical NER as an example, as depicted in Figure~\ref{fig:pre_llm}, we employ the template:
\textit{Please recognize the medical name entity in this sentence: ``[Medical Text]}'', where ``[Medical Text]'' serves as a placeholder for the raw medical text $I_M$.
\textbf{Output Modification}: Instead of using plain targets, we format LLMs outputs into linguistic texts. For the NER task, the recognized head entity
$O_{head}$ and tail entity $O_{tail}$ are integrated into the template:
``\textit{The medical text has the following pairs of entities: head entity is [head entity] and tail entity is [tail entity]}''.
For ease of reference, we label the input and output instruction templates for NER as $TP^Q_{NER}$ and $TP^A_{NER}$, respectively.
With these modifications in place, the process by which LLMs undertake the NER task can be described as follows:
\begin{equation}
    I_M  \rightarrow TP^{Q}_{NER}(I_M) \stackrel{LLM}{\longrightarrow} TP^{A}_{NER}(O_{head}, O_{tail}) \rightarrow O_{head}, O_{tail}
\end{equation}
After the task reformulation for LLMs, we can use the purely lingual data to fine-tune the foundation large language models, such as LlaMA~\cite{touvron2023llama}, ChatGLM~\cite{du2022glm} and etc. The fine-tuned model completes the medical tasks by generating the regulated answers.

\subsection{Multi-task Fine-tuning}
As previously mentioned, medical applications often encompass a variety of tasks, such as name entity recognition, medical inquiry, etc.
Our goal is to fine-tune LLMs to gain robust performance for each task and thus can also benefit the whole healthcare system. 
For multi-task fine-tuning, we consider a set of medical tasks represented as $\mathbb{T}=\{\mathcal{T}_1,\ldots,\mathcal{T}_j,\ldots,\mathcal{T}_M\}$. 
The structured data corresponding to task $\mathcal{T}_j$ can be represented as $\mathcal{D}_{j}=\{(LI_{k}^{\mathcal{T}_j},~LO_{k}^{\mathcal{T}_j})\}^{|\mathcal{D}_j|}_{k=1}$, where $LI$ and $LO$ represent the template-formatted linguistic input and output, respectively.
For the sake of brevity, we omit the superscript $\mathcal{T}_j$ in subsequent discussions.
Assuming the parameters of LLMs are represented by $\Phi$, the multi-task fine-tuning challenge can be articulated as: Given the dataset of all medical tasks $\mathcal{D}=\{\mathcal{D}_j\}^M_{j=1}$, optimize the parameters $\Phi$ of LLMs to ensure optimal performance across each task $\mathcal{T}_j$. 

Since the data from diverse tasks are standardized into a consistent linguistic format, we straightforwardly employ the conditional language modeling objectives~\cite{du2022glm} for all training instances. 
Furthermore, with the intent to assimilate shared medical knowledge and be free from the labor of fine-tuning for various tasks separately, data from all tasks are incorporated into the unique optimization process. 
Consequently, the objective function for multi-task fine-tuning can be formulated as follows:
\begin{equation} \label{eq1}
    \max_{\Phi} \sum_{j\in [M]} \sum_{(x,y)\in \mathcal{D}_j} \sum_{t=1}^{|y|} log(P_{\Phi}(y_t|x,y_{\leq t}))
\end{equation}

\section{Method}
In this section, we provide a comprehensive description of our proposed framework. We begin with an overview of the proposed method. Then, the devised MOELoRA and task-motivated gate are addressed. Finally, we detail fine-tuning and inference processes. %We begin with an overview in Section~\ref{sec:method_overview}. Subsequently, Section~\ref{sec:method_moelora} delves into how MOELoRA seamlessly integrates the processes of MOE and LoRA, harnessing the strengths of both. The task-motivated gate function is detailed in Section~\ref{sec:method_gate}, where we also discuss the recovery of unique fine-tuned LLM parameters for each task. Lastly, Section~\ref{sec:method_opt} elaborates on the optimization and inference processes.

\subsection{Overview} \label{sec:method_overview}
% % %% 双栏图
% %%%%% Framework %%%%%
% \begin{figure}[!t]
% \centering
% \includegraphics[width=1\linewidth]{figures/Method_framework_v4.pdf}
% %\vspace{-5mm}
% \caption{The overview of parameter efficient fine-tuning using MOELoRA.}
% \label{fig:framework}
% \vspace{-5mm}
% \end{figure}
% %%%%% Framework %%%%%

%%%%% Framework %%%%%
\begin{figure*}[!t]
\centering
\includegraphics[width=0.85\linewidth]{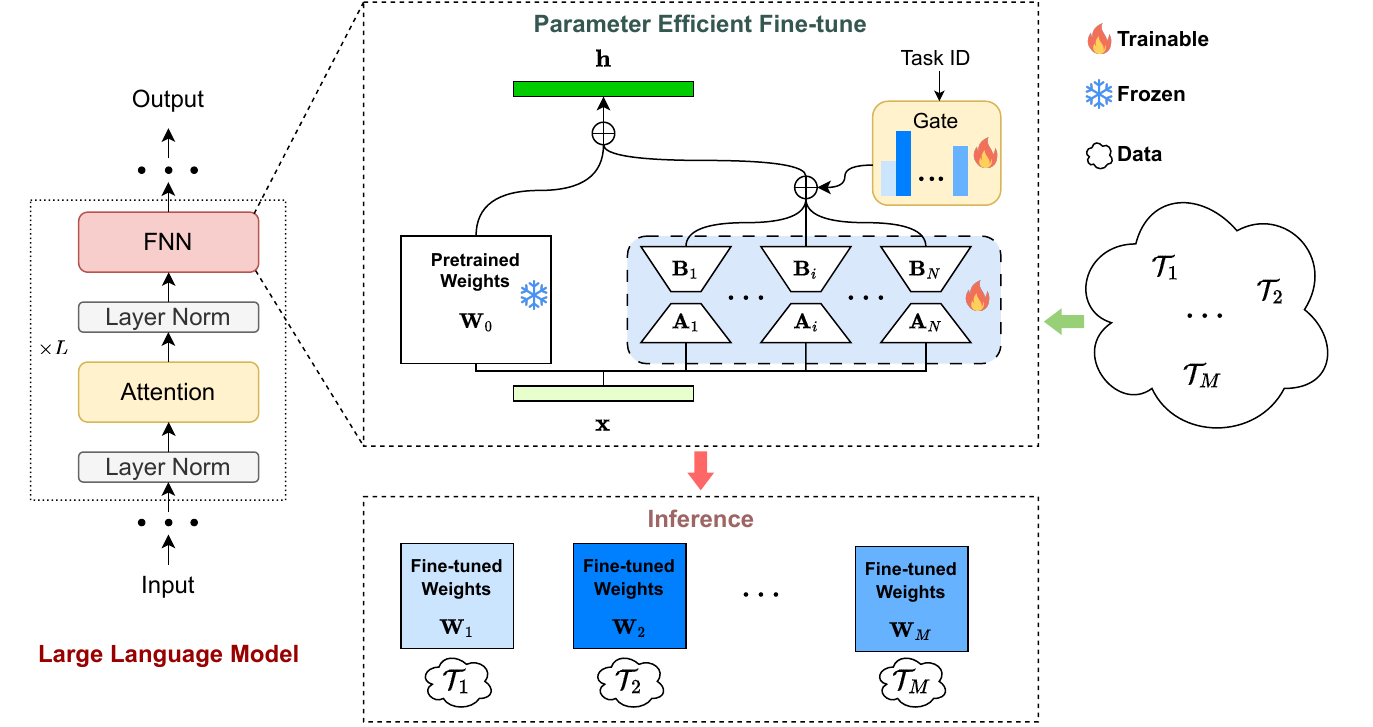}
%\vspace{-5mm}
\caption{The overview of parameter efficient fine-tuning and inference process using MOELoRA.}
\label{fig:framework}
\vspace{-5mm}
\end{figure*}
%%%%% Framework %%%%%
 
Figure~\ref{fig:framework} provides a visual representation of the parameter efficient fine-tuning and inference process of LLMs using MOELoRA. In the realm of parameter efficient fine-tuning, LoRA~\cite{hu2021lora} introduces the concept of training only two low-rank matrices as a substitute for updates in dense layers. Building on this, our approach integrates MOELoRA layers into each dense layer, enabling them to acquire \textit{keys}, \textit{queries}, and \textit{values}, as well as facilitating the feed-forward network (\textit{FNN}). In Figure~\ref{fig:framework}, we take \textit{FNN} as the example for illustration. A significant advantage of our method is that we only fine-tune the parameters of the MOELoRA layers for various tasks, keeping the rest parameters of the original LLMs frozen. 
% This approach substantially reduces the often prohibitive costs associated with tuning.

Furthermore, each MOELoRA layer incorporates multiple experts, which are designed to capture diverse knowledge across various medical tasks, a concept we will delve deeper into in Section~\ref{sec:method_moelora}. Then, we introduce a task-motivated gate function to ensure that unique parameter sets are learned for each task. This function determines the contribution weights of experts across all MOELoRA layers, enabling the generation of distinct updated parameters tailored to different tasks. In particular, we employ a single gate function for all MOELoRA layers, rather than having a one-to-one correspondence between gates and MOELoRA layers. For the fine-tuning process, we update the MOELoRA layers on the mixture of the data from all tasks. Then, the MOELoRA can derive distinct fine-tuned weights for each task during the inference.

% %%%%% MOELoRA %%%%%
% \begin{figure}[!t]
% \centering
% \includegraphics[width=1\linewidth]{figures/Method_MOELoRA_v2.pdf}
% %\vspace{-5mm}
% \caption{The architecture of the proposed MOELoRA.}
% \label{fig:moelora}
% \vspace{-5mm}
% \end{figure}
% %%%%% MOELoRA %%%%%

\subsection{MOELoRA} \label{sec:method_moelora}
Low-rank Adaptation (LoRA)~\cite{hu2021lora} has demonstrated both its effectiveness and efficiency in fine-tuning LLMs. It is inspired by the low intrinsic dimension characteristic~\cite{aghajanyan2021intrinsic}, which reformulates the parameter fine-tuning process in LLMs as a low-rank decomposition. Specifically, the equation $\mathbf{W}_0+\Delta \mathbf{W}=\mathbf{W}+\mathbf{B}\mathbf{A}$ captures this decomposition. Here, $\mathbf{W}_0\in \mathbb{R}^{d_{in} \times d_{out}}$ represents the parameter matrix of the pre-trained LLMs, while $\Delta \mathbf{W}\in \mathbb{R}^{d_{in} \times d_{out}}$ denotes the matrix updated during fine-tuning. The matrices $\mathbf{B} \in \mathbb{R}^{d_{in} \times r}$ and $\mathbf{A} \in \mathbb{R}^{r \times d_{out}}$ are low-rank and trainable. Given this setup, the forward process of a linear layer paired with a LoRA layer can be expressed as:
\begin{equation}    \label{eq2}
    \mathbf{h} = \mathbf{W}_0 \mathbf{x} +  \frac{\alpha}{r} \cdot \Delta \mathbf{W} \mathbf{x} = \mathbf{W}_0 \mathbf{x} + \frac{\alpha}{r} \cdot \mathbf{B} \mathbf{A} \mathbf{x}
\end{equation}

\noindent where $\mathbf{x}$ represents the input vector of dimension $d_{in}$, and $\mathbf{h}$ is the output vector with dimension $d_{out}$. The rank of the trainable low-rank matrices is denoted by $r$, which determines the number of trainable parameters. The constant hyper-parameter $\alpha$ facilitates the tuning of rank $r$~\cite{hu2021lora}. During the LoRA fine-tuning process, all parameters in LLMs remain frozen. Only the low-rank matrices, $\mathbf{A}$ and $\mathbf{B}$, undergo fine-tuning. Given that $r \ll d_{in}$ and $r \ll d_{out}$, the combined number of parameters in $\mathbf{A}$ and $\mathbf{B}$ is significantly smaller than the ones in $\mathbf{W_0}$. Such characteristics result in achieving parameter efficiency for the fine-tuning process.

However, the integral parameters are fine-tuned for all tasks in the original LoRA, which causes difficulty in learning the various aspects of medical knowledge. 
%A potential solution is to derive distinct parameters for various tasks.
A potential solution is to segment the entire parameter set into several parts and derive various combinations for various tasks. The Mixture-of-Expert (MOE) model~\cite{shazeer2017outrageously} suggests employing multiple expert networks to capture different facets of multi-task information, aligning with the combination concept. This insight leads us to design MOELoRA, which seamlessly integrates the advantages of both LoRA and MOE. To harmonize the distinct forward processes of LoRA and MOE, we introduce a set of experts, denoted as $\{E_i\}_{i=1}^N$, to learn the updated matrix $\Delta \mathbf{W}$. As MOELoRA fine-tunes the experts using data from all tasks, it inherently captures shared task knowledge. Moreover, to maintain a compact parameter size, every expert in the MOELoRA layer is constructed as two decomposed low-rank matrices. Based on this structure, the forward process of a linear layer paired with a MOELoRA layer for samples from task $\mathcal{T}_j$ is expressed as:
\begin{equation} \label{eq3}
    \begin{aligned}
        \mathbf{h}_j &= \mathbf{W}_0 \mathbf{x}_j +  \frac{\alpha}{r} \cdot \Delta \mathbf{W}_j \mathbf{x}_j \\
        &=  \mathbf{W}_0 \mathbf{x}_j + \frac{\alpha}{r} \cdot \sum_{i=1}^N \omega_{ji} \cdot  E_i(\mathbf{x}_j) \\
        &= \mathbf{W}_0 \mathbf{x}_j + \frac{\alpha}{r} \cdot \sum_{i=1}^N \omega_{ji} \cdot  \mathbf{B}_i \mathbf{A}_i \mathbf{x}_j
    \end{aligned}
\end{equation}
where $\mathbf{h}_j$ and $\mathbf{x}_j$ represent the input and output of intermediate LLM layers for samples from $\mathcal{T}_j$. 
The matrices $\mathbf{B}_i \in \mathbb{R}^{d_{in} \times \frac{r}{N}}$ and $\mathbf{A}_i \in \mathbb{R}^{\frac{r}{N} \times d_{out}}$ form the expert $E_i$. 
The hyper-parameter $N$ denotes the number of experts in MOELoRA, and for each expert, the rank of matrices $A$ and $B$ is $\frac{r}{N}$. 
% To ensure that distinct parameters are learned for different tasks, the contribution of each expert should be task-specific. 
In Equation~\eqref{eq3}, the term $\omega_{ji}$ modulates these contribution weights for task $\mathcal{T}_j$. This weight is determined by our proposed gate function, which will be detailed later.

Here, we will discuss the number of trainable parameters for LoRA and MOELoRA. In terms of LoRA, the two low-rank matrices $\mathbf{B} \in \mathbb{R}^{d_{in} \times r}$ and $\mathbf{A} \in \mathbb{R}^{r \times d_{out}}$ contain all trainable parameters. Thus, the number of trainbale parameters of LoRA is $d_{in} \times r+r \times d_{out}=r \times (d_{in}+d_{out})$. As for MOELoRA, there are $N$ trainable experts and each expert owns $\frac{r}{N} \times (d_{in}+d_{out})$, so total number is calculated as $N \times \frac{r}{N} \times (d_{in}+d_{out})=r \times (d_{in}+d_{out})$. In conclusion, MOELoRA has the same number of trainable parameters as LoRA, which indicates high efficiency.

\subsection{Task-Motivated Gate Function} \label{sec:method_gate}
In this section, we detail the intricacies of our task-motivated gate function. As previously emphasized, the contribution of each expert should be tailored to specific tasks. To regulate these contributions, we introduce a gate function. Since these weights are inherently task-specific, our gate function is designed to take the task identity as input. To facilitate this, we employ a task embedding matrix, denoted as $\mathbf{E} \in \mathbb{R}^{|\mathbb{T}| \times d_{T}}$, where $d_T$ represents the dimension of the task embedding. Upon identifying a task $\mathcal{T}_j$, we extract the $j$-th column of $\mathbf{E}$, which serves as the representation vector for that task, symbolized as $\mathbf{e}_j \in \mathbb{R}^{d_T}$. To determine the contribution weights for task $\mathcal{T}_j$, we apply a linear transformation. This computation is captured by the following equation:
\begin{equation} \label{eq4}
\bm{\omega}_j = {\rm Softmax}(\mathbf{W}_T \mathbf{e}_j)
\end{equation}

\noindent Here, $\bm{\omega}_j \in \mathbb{R}^{N}$ represents the contribution weight vector tailored for task $\mathcal{T}_j$. The transformation matrix is denoted as $\mathbf{W}_T \in \mathbb{R}^{N \times d_T}$. To prevent any disproportionately large weights, we employ a softmax operation to normalize the contribution weights.

The gate mentioned in Equation~\eqref{eq4} is naturally a dense design to combine all of the experts. Recently, some works~\cite{shazeer2017outrageously,riquelme2021scaling} have focused on another form of gate, \ie sparse gate. It has the benefit of alleviating the optimization interference across various tasks~\cite{gupta2022sparsely}. Thus, we also design a sparse version of the task-motivated gate to explore which design is more effective. The devised sparse gate is formulated as follows:
\begin{equation}
    \bm{\omega}_j = {\rm Softmax}({\rm Top}(\mathbf{W}_T \mathbf{e}_j, K))
\end{equation}
\begin{equation}
    {\rm Top}(\mathbf{x}, K)=
    \begin{cases}
        x_i & {\rm if} \ x_i \ {\rm in} \ {\rm top} \ K \  {\rm elements} \ {\rm of} \ \mathbf{x}. \\
        0 & \rm{otherwise.}
    \end{cases}
\end{equation}

% Next, we introduce how the proposed gate function derives distinct parameters for each task. 
While the conventional design of MOE directly feeds the input vector $\mathbf{x}$ into the gate function, our approach diverges. We exclusively input the task identity into the gate function, as Figure~\ref{fig:framework} shows, aiming to yield a unique set of model parameters for each task. To illustrate, if one wishes to recover the fine-tuned parameters for task $\mathcal{T}_j$, the process can be articulated as:
\begin{equation}    \label{eq5}
\begin{aligned}
    \mathbf{W}_j &= \mathbf{W}_0 +  \frac{\alpha}{r} \cdot \Delta \mathbf{W}_j \\
    &=  \mathbf{W}_0 + \frac{\alpha}{r} \cdot \sum_{i=1}^N \omega_{ji} \cdot  E_i \\
    &= \mathbf{W}_0 + \frac{\alpha}{r} \cdot \sum_{i=1}^N \omega_{ji} \cdot  \mathbf{B}_i \mathbf{A}_i 
\end{aligned}
\end{equation}

\noindent If the gate function is driven by the input vector $\mathbf{x}$, the weight vector would differ across samples. This means each sample would possess its unique $\bm{\omega}_j$, leading to a sample-specific fine-tuned parameter matrix. This design would render the parameters non-recoverable on a per-task basis. The ability to recover parameters for each task offers two primary advantages:
1) \textit{Customization for Task}: Each task is fine-tuned with a set of parameters, which can help learn more task-specific information and alleviate the problem of data imbalance.
2) \textit{Efficiency in Inference}: The recovered, fine-tuned LLMs exhibit reduced inference latency. This is attributed to the elimination of the need for the additional forward computation associated with the MOELoRA layer.
% 2) Deployment Flexibility: It facilitates the deployment of the fine-tuned LLM tailored for individual tasks.

\subsection{Fine-tune and Inference} \label{sec:method_opt}
In this section, we refer to the fine-tuning and inference process of MOELoRA. For better readability, we also conclude the whole procedure in Algorithm~\ref{alg:opt}. 

\textbf{Fine-tuning}. We first configure the MOELoRA according to the specified layers in LLMs and several hyper-parameters (line 1-3). Then, for the parameter efficient fine-tuning, all pre-trained parameters in LLMs (line 4) are frozen. During the fine-tuning, we randomly sample a batch of data from all tasks iteratively, instead of grouping the samples from the same task into one batch as some multi-task researches~\cite{sheng2021one,sun2023multitask} do. We choose the random sampling for batch by the performance comparison in experiments. Using the batch of data, we can conduct the forward process and compute the loss for fine-tuning (line 6-7). For parameter update, it is worth noting that we only fine-tune the parameters of MOELoRA and task-motivated gate function, \ie $\{\mathbf{A}_i,\mathbf{B}_i\}_{i=1}^N$ and $\{\mathbf{E}, \mathbf{W}_T\}$.

\textbf{Inference}. As discussed in Section~\ref{sec:method_gate}, the MOELoRA can recover the fine-tuned parameter matrices for each task by Equation~\eqref{eq5}. For inference, we first recover the fine-tuned parameters in LLMs for each task (line 10-13), which indicates that each task has its own LLMs parameters. Then, we can apply the corresponding LLMs to complete the specified task.

%% 算法步骤
\let\oldnl\nl% Store \nl in \oldnl
\newcommand{\nonl}{\renewcommand{\nl}{\let\nl\oldnl}}% Remove line number for one line
%%%%% MOELoRA Algorithm %%%%%
\begin{algorithm}[t]
\caption{Fine-tuning and inference process of MOELoRA} \label{alg:opt}
%\LinesNumbered
\raggedright

\begin{algorithmic} [1]
    \State Indicate the LLMs and the layers that need MOELoRA fine-tuning.
    \State Indicate the rank value $r$ and scale value $\alpha$.
    \State Indicate the number of experts $N$ of MOELoRA.
\end{algorithmic}

\textbf{Fine-tuning Process} 
\setcounter{algorithm}{2}
\begin{algorithmic} [1]
    \makeatletter
    \setcounter{ALG@line}{3}
    \State Freeze all parameters in pre-trained LLMs, \eg $\mathbf{W}_q$, $\mathbf{W}_k$, $\mathbf{W}_v$.
    \For {a batch of samples $B$ in $\mathcal{D}$}
        \State Conduct forward process for LLMs accompanied with MOELoRA by Equation~\eqref{eq3}.
        \State Compute the loss function by Equation~\eqref{eq1}.
        \State Update the parameters of MOELoRA $\{\mathbf{A}_i,\mathbf{B}_i\}_{i=1}^N$ and the parameters of gate function $\{\mathbf{E}, \mathbf{W}_T\}$
    \EndFor
\end{algorithmic}

\textbf{Inference Process}
\setcounter{algorithm}{8}
\begin{algorithmic} [1]
    \makeatletter
    \setcounter{ALG@line}{9}
    \For {$\mathcal{T}_j$ in $\mathbb{T}$}
        \State Calculate the contribution weights $\bm{\omega}_j$ for each experts by Equation~\eqref{eq4}.
        \State Recover the MOELoRA fine-tuned parameters by Equation~\eqref{eq5} for each task.
    \EndFor
    \State For specific task $\mathcal{T}_j$, apply the corresponding parameters of LLMs for prediction.
\end{algorithmic}
\end{algorithm}
%%%%% MOELoRA Algorithm %%%%%

\section{Experiment}

In this section, we seek to address the following Research Questions (\textbf{RQ}):

\begin{itemize}[leftmargin=*]
    \item \textbf{RQ1}: How does MOELoRA compare to other parameter-efficient fine-tuning strategies and cross-task generalization methods in terms of performance?
    
    \item \textbf{RQ2}: What impact do the MOE architecture and the gate function have on the fine-tuning process? How do different training strategies influence MOELoRA's performance?
    
    \item \textbf{RQ3}: How do the number of experts and the rank of MOELoRA influence performance outcomes?

    \item \textbf{RQ4}: Is the proposed MOELoRA efficient in the process of fine-tuning and inference?
    
    \item \textbf{RQ5}: Are the experts specialized in capturing specific aspects of knowledge for various tasks?
\end{itemize}

\subsection{Experimental Settings}

\subsubsection{\textbf{Dataset}}
Our experiments are conducted on the PromptCBLUE dataset\footnote{\url{https://tianchi.aliyun.com/competition/entrance/532084/information}}~\cite{zhu2023promptcblue}, a multi-task Chinese medical dataset recently made available on the Tianchi Competition Platform\footnote{\url{https://tianchi.aliyun.com/competition/activeList}}. 
This dataset encompasses $16$ distinct medical tasks, each of which has been transformed into pure text format using specific prompts, ensuring compatibility with LLMs. 
% For each task, multiple templates actually exist and are used in the experiments for a better generalization~\cite{zhang2023instruction}. 
To the best of our knowledge, PromptCBLUE is the only multi-task medical dataset tailored for LLMs. 
% Specifically, the dataset includes tasks such as medical named entity recognition, diagnosis report generation, etc. 
Due to computational constraints, we have chosen $8$ tasks at random for our experiments. For pre-processing, we eliminate duplicate samples in the original dataset. 
Since the test set used in the competition remains unreleased, we opt to use the development set as our test set. Then, the validation set for the experiment is derived from the training set in competition, with its size matching that of the test set. 
% It is worth noting that the split validation set is not used for fine-tuning. 
The statistics of the dataset are concluded in Table~\ref{tab:exp_dataset}. 

%%% Statistics of Datasets %%%
\begin{table}[!t]
\centering
\caption{The brief description and statistics of the dataset PromptCBLUE.}
\resizebox{0.47\textwidth}{!}{
\begin{tabular}{llccc}
\toprule[1.5pt]
Task & Description & \# Train & \# Validation & \# Test \\ 
\midrule
CMeIE & Name Entity Recognition & 2,828 & 600 & 600 \\
CHIP-CDN & Normalization & 2,381 & 600 & 600 \\
CHIP-CDEE & Attribute Extraction & 1,562 & 600 & 600 \\ 
CHIP-MDCFNPC & Clinic Entity Discovery & 4,935 & 600 & 600 \\ 
CHIP-CTC & Medical Text Classification & 3,622 & 1,100 & 1,100 \\ 
KUAKE-QIC & Query Intention & 3,279 & 660 & 660 \\ 
IMCS-V2-MRG & Report Generation & 1,799 & 600 & 600 \\ 
MedDG & Doctor Dialogue & 4,964 & 600 & 600 \\ 
\bottomrule[1.5pt]
\end{tabular}
}
\label{tab:exp_dataset}
\vspace{-5mm}
\end{table}
%%% Statistics of Datasets %%%

\subsubsection{\textbf{Baselines}}
In our experiments, we benchmark against four distinct groups of baselines:
%, namely: (i) LLM without Fine-tuning, (ii) LLM with Fine-tuning, (iii) Model Editing, and (iv) Cross-task Generalization. The latter three groups utilize ChatGLM-6B~\cite{du2022glm}, renowned for its prowess in Chinese text generation. 
% A brief description of each baseline group is as follows.

\noindent \textbf{LLMs without Fine-tuning}: We employ In-Context Learning~\cite{dong2022survey} to guide LLMs in accomplishing the relevant tasks. 
\begin{itemize}[leftmargin=*]
    \item \textbf{ChatGPT}~\cite{brown2020language}. ChatGPT is one of the most popular LLMs. 
    % It is pre-trained on a large corpus of various data, which imbues it with the potential to complete medical tasks. 
    To inspire task-relevant ability, we randomly sample $3$ to $10$ input-output pairs from the training data of one specific task to get the demonstration, filling the input length to the maximum.

    \item \textbf{Huatuo}~\cite{wang2023huatuo}. Huatuo first constructs a Chinese medical instruction dataset by a medical database. In our experiment, we use the version of ChatGLM-6B for fair comparison and the same in-context learning method as the ChatGPT baseline.
\end{itemize}

\noindent \textbf{LLMs with Fine-tuning}: This group encompasses various strategies of the fine-tuning. 
\begin{itemize}[leftmargin=*]
    \item \textbf{P-Tuning}~\cite{liu2023gpt}. \textit{P-Tuning} designs a trainable prompt encoder to produce continuous prompt vectors, which are inserted into the input sequence. We implement it by fine-tuning the prompt encoder on the data of all tasks.

    \item \textbf{LoRA (Full)}~\cite{hu2021lora}. LoRA designs two low-rank matrices as the trainable parameters for dense layers while freezing all parameters of pre-trained LLMs. \textit{LoRA (Full)} trains a unique set of LoRA parameters for all tasks.

    \item \textbf{LoRA (Single)}~\cite{hu2021lora}. We implement \textit{LoRA (Single)} by separately training LoRA for each task. For the time and resource limitation, we adopt the same set of hyper-parameters for all tasks and select the model according to the best average score.

    \item \textbf{LoRA (Full+TP)}~\cite{hu2021lora}. We add simple task demonstration to input texts, which aims to prompt LLMs with the distinctions between tasks. As for the implementation, we conduct the same training process as \textit{LoRA (Full)}.
\end{itemize}

\noindent \textbf{Model Editing}: One model editing method can be adapted to multi-task fine-tuning for LLMs.
\begin{itemize}[leftmargin=*]
    \item \textbf{Task-Arithmetic}~\cite{ilharco2022editing}. \textit{Task-Arithmetic} defines a novel task vector, which can be operated for merging or unlearning tasks. However, it is for full fine-tuning, so we modify the method for fair comparison: the task vector is calculated by $\tau_t=\textbf{B}_t \cdot \textbf{A}_t$, where $\textbf{A}$ and $\textbf{B}$ are the low-rank matrices in LoRA layers. According to the adding operation for multi-task, we add all task vectors together and adjust the scale factor on the validation set.
\end{itemize}

\noindent \textbf{Cross-task Generalization}: To assess the applicability of cross-task generalization to multi-task fine-tuning, we also evaluate two recent approaches, \textit{LoRAHub}~\cite{huang2023lorahub} and \textit{MoLoRA}~\cite{zadouri2023pushing}.
\begin{itemize}[leftmargin=*]
    \item \textbf{LoRAHub}~\cite{huang2023lorahub}. \textit{LoRAHub} proposes an assembling method to compose LoRA parameters fine-tuned on source tasks and seek the generalization to unseen target tasks. In the experiment, we LoRA fine-tune each task. Then, use the validation of one specified task to learn the composing weight and test the performance for this task. 
    \item \textbf{MoLoRA}~\cite{zadouri2023pushing}. \textit{MoLoRA} is a relatively recent work, which adopts the MOE structure to the LoRA. However, the gate in MoLoRA takes the intermediate embedding of the tokens to derive expert weights. In our experiments, we adapt it to our multi-task setting, \ie training and testing on same set of tasks.
\end{itemize}

\subsubsection{\textbf{Implementation Details}} 
Our experiments are simulated by PyTorch 1.12.0 and Python 3.9.5. We run the code on Tesla V100 32G GPUs for acceleration. The LLM ChatGLM-6B~\cite{du2022glm}, recognized for its proficiency in Chinese language processing, serves as the foundational model for fine-tuning. For all LoRA fine-tuning baselines and the proposed MOELoRA, we designate trainable layers for the layers identified as ``query\_key\_value'', ``dense'', ``dense\_h\_to\_4h'', and ``dense\_4h\_to\_h''. The maximum input and output lengths were configured to $1,024$ and $196$, respectively. We set the batch size to $64$ with a maximum of $8,000$ training steps. The LoRA rank $r$ was fixed at $16$, with a LoRA dropout $\alpha = 0.1$. For MOELoRA, the number of experts is set to $8$. $K$ is searched from $1$ to $7$ for the sparse gate version of MOELoRA, and we find $2$ is the optimal.
% Since the LoRA fine-tuning for ChatGLM-6B is still highly time-consuming, we adopt parallel computation. In detail, 
% Besides, we utilize the deepspeed package~\cite{rajbhandari2020zero} to distribute the computation on several Tesla V100 32G GPUs and conduct gradient accumulation for the limited GPU memory. To balance the time and accuracy, we choose the half-precision training, denoted as FP16, for all experiments. 
Besides, during the testing, we set the temperature as $0.95$ for generation. Our MOELoRA implementation\footnote{\url{https://github.com/Applied-Machine-Learning-Lab/MOELoRA-peft}} is compatible with the PEFT package\footnote{\url{https://github.com/huggingface/peft}}, which can facilitate easier adoption and utilization of the proposed MOELoRA.

\subsubsection{\textbf{Evaluation Metrics}} 
For our evaluations, we employ a variety of metrics tailored to the nature of each task. For example, CMeIE is a task of name entity recognition (NER), where there are too many entity classes (\textbf{1262} classes for CMeIE), so we apply commonly used Micro-F1 for this task~\cite{wang2022nested}. CHIP-CDN (\textbf{579} classes), CHIP-CDEE (\textbf{998} classes) and CHIP-MDCFNPC (\textbf{2065} classes) are all tasks that own too many categories, so Micro-F1 is used to evaluate them too. By comparison, CHIP-CTC (\textbf{44} classes) and QUAKE-QIC (\textbf{7} classes) tasks both have fewer classes, which requires considering the equal importance of each class~\cite{opitz2019macro}, so we apply Macro-F1 for them. As for text generation tasks, \ie IMCS-V2-MRG and MedDG, the Rouge-L~\cite{lin2004automatic} is applied. Also, the average score across all tasks is used for evaluating overall performance. To ensure the robustness and reproducibility of our results, tests are run thrice by random seeds $\{42,43,44\}$, with average scores reported in the following experimental results.

%%% Overall Performance %%%
\begin{table*}[!t]
\centering
\scriptsize
\tabcolsep=0.08cm   % 调整列间距
\caption{The overall results of competing baselines and MOELoRA on PromptCBLUE. The boldface refers to the highest score and the underline indicates the best result of the methods. ``\textbf{{\Large *}}'' indicates the statistically significant improvements (\ie two-sided t-test with $p<0.05$) over the best baseline.}
\resizebox{1\textwidth}{!}{
\begin{tabular}{l|cccccccc|c}
\toprule[1pt]
\textbf{Model} & \textbf{CMeIE} & \textbf{CHIP-CDN} & \textbf{CHIP-CDEE} & \textbf{CHIP-MDCFNPC} & \textbf{CHIP-CTC} & \textbf{KUAKE-QIC} & \textbf{IMCS-V2-MRG} & \textbf{MedDG} & \textbf{Avg.}   \\ 
\midrule
% Bert &  &   &   &   &   &   &   &   &        \\ 
% \midrule
ChatGPT & 0.3058 & 0.6069 & 0.2838 & 0.5854 & 0.5253 & 0.4851 & 0.3253 & \textbf{0.1361} & 0.4067 \\
Huatuo & 0.1826 & 0.3610 & 0.1658 & 0.3487 & 0.1909  & 0.1454 & 0.2401  & \underline{0.1308}  & 0.2207      \\ 
\midrule
P-Tuning & 0.4552 & 0.8687 & 0.5256 & 0.7423 & 0.8275 & 0.8377 & 0.3155 & 0.0901 & 0.5828 \\
LoRA (Full) & 0.5089 & 0.8748 & 0.5464 & 0.7780 & \textbf{0.8758} & 0.8615 & \underline{0.3678} & 0.1113 & 0.6155 \\
LoRA (Single) & 0.4984 & 0.8882 & 0.5528 & 0.7765 & 0.8694 & 0.8524 & 0.3583 & 0.1143 & 0.6138 \\
LoRA (Full+TP) & 0.4933 & 0.8814 & 0.5450 & 0.7705 & \underline{0.8755} & \underline{0.8664} & 0.3556 & 0.1160 & 0.6130 \\ 
\midrule
Task-Arithmetic & 0.3928 & 0.7533 & 0.3216 & 0.6619 & 0.8091 & 0.6596 & 0.3163 & 0.1147 & 0.5037 \\ 
\midrule
LoRAHub & 0.4411 & 0.8442 & 0.5041 & 0.7177 & 0.8564 & 0.8502 & 0.3061 & 0.1192 & 0.5799 \\ 
MoLoRA & 0.5081 & 0.8850 & 0.5656 & 0.7850 & 0.8749 & 0.8605 & 0.3590 & 0.1067 & 0.6181 \\
\midrule
\textbf{MOELoRA(D)} & \textbf{0.5193}* & \underline{0.8928}* & \underline{0.5697}* & \textbf{0.7933}* & 0.8691 & \textbf{0.8675} & \textbf{0.3681} & 0.1089 & \textbf{0.6236}* \\ 
\textbf{MOELoRA(S)} & \underline{0.5110}* & \textbf{0.8980}* & \textbf{0.5719}* & \underline{0.7872}* & 0.8682 & 0.8633 & 0.3558 & 0.1080 & \underline{0.6204}* \\
\bottomrule[1pt]
\end{tabular}
}
\label{tab:exp_overall}
\vspace{-4mm}
\end{table*}
%%% Overall Performance %%%

\subsection{Overall Performance (RQ1)}
The comprehensive experimental results of MOELoRA and the competing baselines are presented in Table~\ref{tab:exp_overall}. MOELoRA(D) and MOELoRA(S) represent the dense and sparse gate design for the MOELoRA, respectively. Analyzing the average scores across all tasks, it is evident that MOELoRA(D) consistently outperforms all other methods. To respond \textbf{RQ1}, the detailed analysis is given:
\begin{itemize}[leftmargin=*]
    \item \textbf{LLMs without Fine-tuning}: The first group of baselines, which are LLMs without any fine-tuning, significantly lag behind the other groups. This highlights the importance of fine-tuning LLMs to incorporate task-specific medical knowledge. 
    %Notably, ChatGPT outperforms Huatuo on most tasks, suggesting that the LLMs only in large parameter scales can be well motivated by the in-context learning~\cite{dong2022survey}.

    \item \textbf{Parameter Efficient Fine-tuning Strategies}: Among the parameter efficient fine-tuning strategies, LoRA-based methods clearly surpass P-Tuning. 
    While LoRA (Full) and LoRA (Full+TP) both utilize data from all tasks, LoRA (Full+TP) slightly underperforms. This might be attributed to the addition of task prompts, which extend the input texts, leading to the potential truncation of informative words due to input length constraints. LoRA (Single), which fine-tunes for individual tasks, also does not match the performance of LoRA (Full), underscoring the value of shared knowledge across tasks.

    \item \textbf{Model Editing}: The Task-Arithmetic evidently underperforms all tuning competitors. The reason may be that the model editing method is not suitable to the parameter efficient fine-tuning.
    
    \item \textbf{Cross-task Generalization}: We benchmark against two recent cross-task generalization methods. Despite their impressive performance in cross-task generalization settings, they require a vast amount of task data, which conflicts with the multi-task setting. In our experiments, we only consider 8 tasks, which might explain its relative underperformance. 
    
    \item \textbf{Input-related Gate V.S. Task-related Gate}: MoLoRA can be considered as an input-related gate variant of our MOELoRA. We can find that MOELoRA outperforms MoLoRA. The reason could lie in that MoLoRA needs different gate functions for each MoLoRA layer, which leads to many redundant parameters and difficulty in training. Such performance comparison validates the effectiveness of task-motivated gate design.

    \item \textbf{Dense Gate V.S. Sparse Gate}: From the Table~\ref{tab:exp_overall}, we find that MOELoRA(S) performs the best on two tasks. The reason lies in that the sparse gate can help alleviate the optimization interference across various tasks~\cite{gupta2022sparsely}, which can help learn some specific tasks better. However, for multi-task medical applications, shared medical knowledge is more vital. The dense gate can benefit the model learning shared knowledge by utilizing all experts, so such a design shows a superior performance on most tasks.

    \item \textbf{Task-specific Observations}: Performance variations are evident across tasks. For instance, LoRA (Full) and LoRA (Full+TP) excel in tasks with larger datasets, while LoRA (Single) shines in tasks with fewer samples, highlighting the data imbalance issue. MOELoRA consistently achieves optimal performance in most tasks, demonstrating its ability to effectively address this imbalance. For MedDG tasks, the inherent dialog capability of ChatGPT and Huatuo gives them an advantage over other approaches.   
    
\end{itemize}
% In response to \textbf{RQ1}, MOELoRA demonstrates superior performance compared to other parameter-efficient strategies and cross-task generalization methods.

%%% Abltion Study %%%
\begin{table*}[!t]
\centering
\tabcolsep=0.08cm   % 调整列间距
\scriptsize
\caption{The experimental results of ablation study for MOELoRA. The boldface refers to the highest score and the underline indicates the best result of the methods. ``\textbf{{\Large *}}'' indicates the statistically significant improvements (\ie two-sided t-test with $p<0.05$) over the best baseline.}
\resizebox{1\textwidth}{!}{
\begin{tabular}{l|cccccccc|c}
\toprule[1pt]
\textbf{Model} & \textbf{CMeIE} & \textbf{CHIP-CDN} & \textbf{CHIP-CDEE} & \textbf{CHIP-MDCFNPC} & \textbf{CHIP-CTC} & \textbf{KUAKE-QIC} & \textbf{IMCS-V2-MRG} & \textbf{MedDG} & \textbf{Avg.}   \\ 
\midrule
\textit{w/o} MOE & \underline{0.5089} & 0.8748 & 0.5464 & 0.7780 & 0.8758 & 0.8615 & \underline{0.3678} & 0.1113 & 0.6155 \\
\textit{w/o} gate  & 0.5015 & \underline{0.8840} & 0.5378 & 0.7789 & \textbf{0.8818} & \underline{0.8699} & 0.3709 & \textbf{0.1140} & 0.6174 \\
\textit{w} multiple gate & 0.4994 & \underline{0.8840} & 0.5692 & 0.7842 & \underline{0.8764} & 0.8675 & 0.3632 & 0.1130 & \underline{0.6196}       \\ 
\midrule
\textit{w} BT   & 0.4817 & 0.8806 & \textbf{0.5712} & 0.7713 & 0.8682 & 0.8643 & 0.3522 & 0.1110 & 0.6126 \\
\textit{w} RBT  & 0.4769 & 0.8830 & 0.5600 & 0.7741 & 0.8636 & \textbf{0.8795} & 0.3541 & \underline{0.1135} & 0.6131 \\
\midrule
LoRA (Full)-QKV & 0.4666 & 0.8605 & 0.4997 & 0.7703 & 0.8264 & 0.7161 & 0.3636 & 0.1123 & 0.5770 \\
MOELoRA(D)-QKV & 0.4897 & 0.8733 & 0.5227 & \underline{0.7854} & 0.8213 & 0.7295 & 0.3640 & 0.1099 & 0.5869 \\ 
\midrule
\textbf{MOELoRA(D)} & \textbf{0.5193}* & \textbf{0.8928}*   & \underline{0.5697}   & \textbf{0.7933}*       & 0.8691   & 0.8675    & \textbf{0.3681}      & 0.1089 & \textbf{0.6236}* \\ 
\bottomrule[1pt]
\end{tabular}
}
\label{tab:exp_ablation}
\vspace{-4mm}
\end{table*}
%%% Ablation Study %%%

\subsection{Ablation Study (RQ2)}
To delve deeper into \textbf{RQ2} and understand the contributions of each component in MOELoRA, we present the results of our ablation study in Table~\ref{tab:exp_ablation}. The variant \textit{w/o MOE} (essentially reverts to LoRA (Full)) excludes the MOE architecture. It demonstrates inferior performance compared to the full-fledged MOELoRA, underscoring the significance of the MOE architecture. Similarly, the \textit{w/o gate} variant, which employs uniform expert weights bypassing the gate function, also lags behind MOELoRA, highlighting the gate function's effectiveness. The \textit{w multiple gate} variant, uses a unique gate function for each MOELoRA layer. We can see that it achieves comparable results on several tasks and is slightly outperformed by the single gate function design due to over-parameterization. Besides, multiple gate functions also incur a higher count of trainable parameters, leading to diminished efficiency compared to a single gate function setup.

Additionally, we analyze the impact of different training strategies. Specifically, the \textit{w BT} method~\cite{sheng2021one} consolidates samples from the same task into one batch. The \textit{w RBT} approach~\cite{sun2023multitask} randomly selects a task for each batch of data. Both of them prove to be less conducive for MOELoRA, resulting in performance degradation. This performance comparison underscores the influence of specific training patterns.
% These findings underscore the critical roles of both the MOE architecture and the gate function in the MOELoRA model, as well as the influence of specific training patterns.

For verification of the robustness of the proposed MOELoRA, We have conducted the experiments of only imposing LoRA layers on the attention layers, denoted as \textit{LoRA (Full)-QKV} and \textit{MOELoRA(D)-QKV}. From the results, we find that \textit{MOELoRA(D)-QKV} can outperform \textit{LoRA (Full)-QKV} on most tasks, which aligns the performance comparison between \textit{MOELoRA(D)} and \textit{LoRA (Full)} in Table~\ref{tab:exp_overall}. Besides, \textit{MOELoRA(D)} is better than \textit{MOELoRA(D)-QKV}, which illustrates more MOELoRA layers can raise the performance of fine-tuning consistently.

%%%% Hyper-parameter Experiments %%%%%
\begin{figure}[!t]
\begin{minipage}[t]{0.49\linewidth}
\centering
\begin{subfigure}{1\linewidth}
    \includegraphics[scale=0.24]{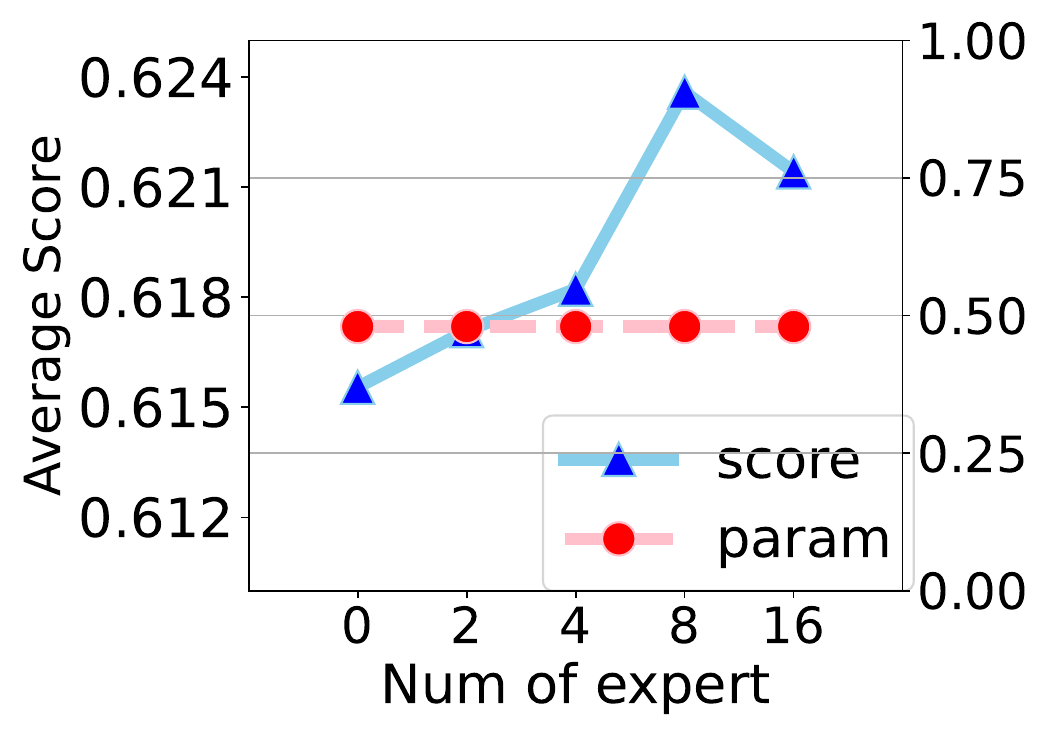}
    \caption{}
    \label{fig:exp_expert}
\end{subfigure}
\end{minipage}%
% \hspace{3mm}
\begin{minipage}[t]{0.49\linewidth}
    \centering
\begin{subfigure}{1\linewidth}
    \includegraphics[scale=0.24]{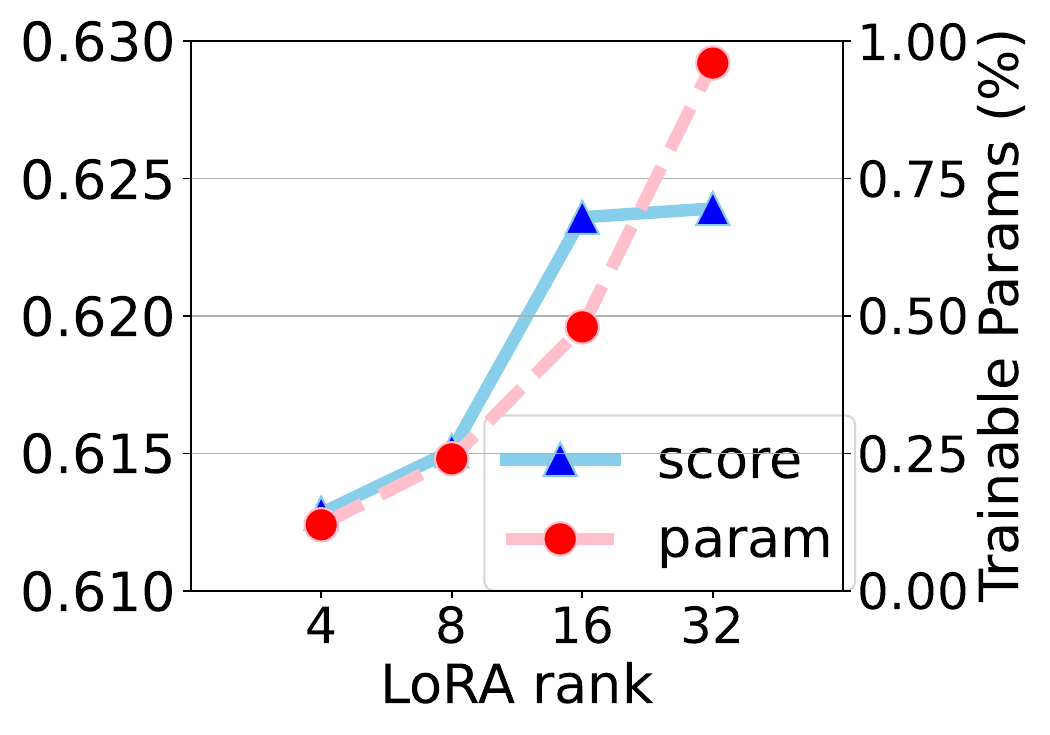}
    \caption{}
    \label{fig:exp_rank}
\end{subfigure}
\end{minipage}%
\caption{The results of experiments for hyper-parameters, \ie expert number $N$ and LoRA rank $r$.}
\label{fig:exp_hyper}
\vspace{-4mm}
\end{figure} 
%%%%% Hyper-parameter Experiments %%%%%

\subsection{Hyper-parameter Analysis (RQ3)}

To answer \textbf{RQ3}, we delve into the impact of hyper-parameters on the performance of MOELoRA(D). 
Specifically, we examine how variations in the expert number $N$ and LoRA rank $r$ influence the results, as depicted in Figure~\ref{fig:exp_hyper}.
Our observations reveal that the performance of MOELoRA improves as $N$ increases from $0$ to $8$, while fixing the overall LoRA rank $r$ as $16$. 
This enhancement can be attributed to the fact that a greater number of experts facilitate the learning of a broader spectrum of knowledge~\cite{shazeer2017outrageously}. 
However, when $N$ escalates to $16$, we notice a marginal decline in performance. 
This can be explained by a large expert number leading to a small LoRA rank for each expert, which degrades the learning ability of low-rank matrices.
% It is worth noting that during these experiments, we maintain a constant overall LoRA rank $r$, ensuring that the size of trainable parameters remains relatively stable. 
Subsequently, we set the rank of each expert, \ie $\frac{r}{N}$, as $2$.
We can observe from Figure~\ref{fig:exp_rank} that while an increase in $r$ consistently boosts performance, it also leads to a proportionate surge in the size of trainable parameters. 
Given the need to strike a balance between efficiency and performance, a practical choice for $r$ would be $16$.

%%%% Efficiency Experiments %%%%%
\begin{figure}[!t]
\begin{minipage}[t]{0.49\linewidth}
\centering
\begin{subfigure}{1\linewidth}
    \includegraphics[scale=0.24]{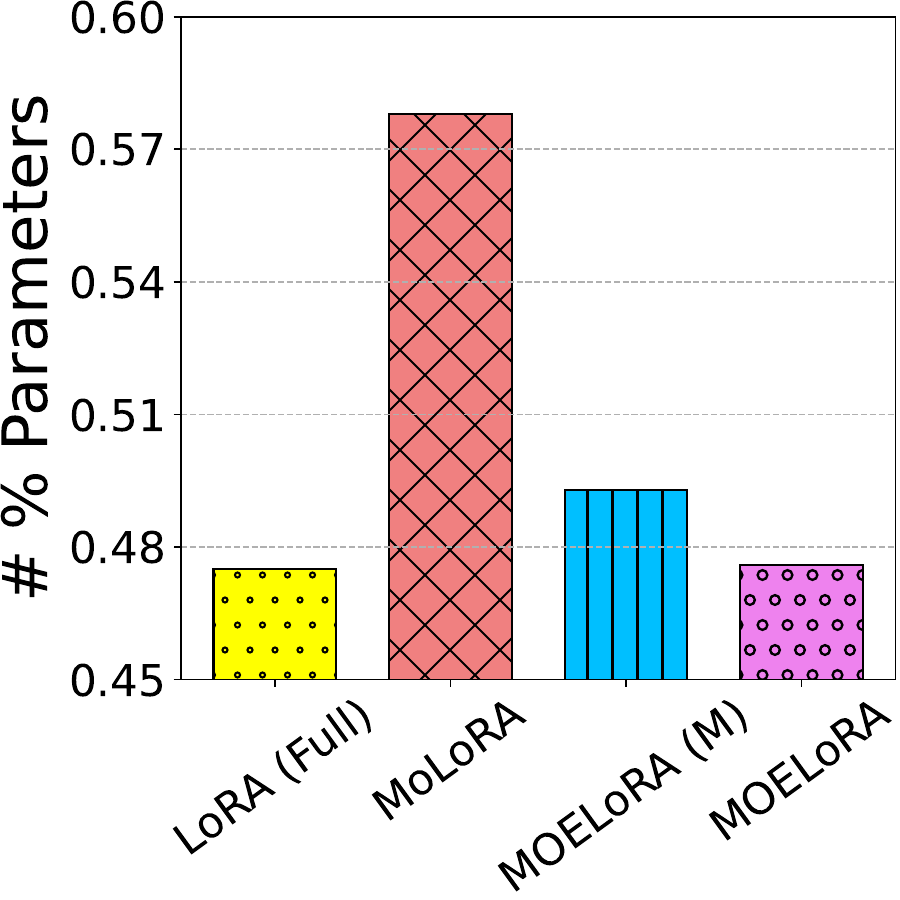}
    \caption{Ratio of Parameters}
    \label{}
\end{subfigure}
\end{minipage}%
% \hspace{3mm}
\begin{minipage}[t]{0.49\linewidth}
    \centering
\begin{subfigure}{1\linewidth}
    \includegraphics[scale=0.24]{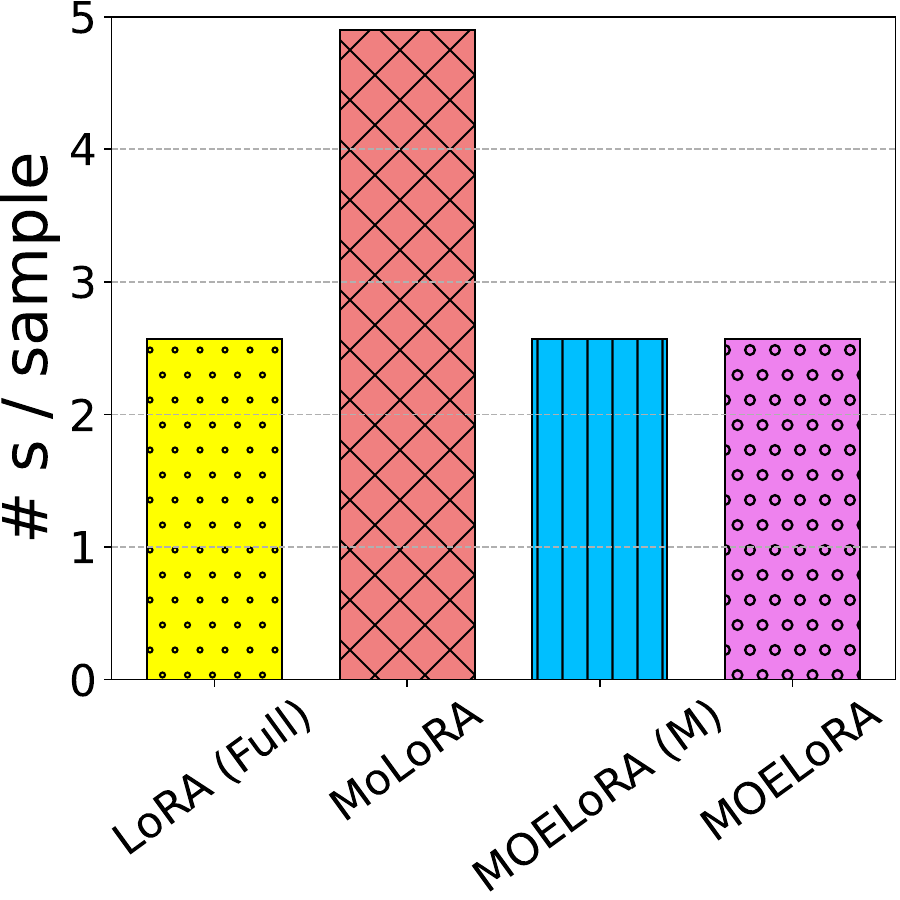}
    \caption{Inference Latency}
    \label{}
\end{subfigure}
\end{minipage}%
\caption{The results of experiments for comparing training and inference efficiency.}
\label{fig:exp_efficiency}
\vspace{-4mm}
\end{figure} 
%%%%% Efficiency Experiments %%%%%

\subsection{Efficiency Analysis (RQ4)}

To evaluate the training and inference efficiency, we compare the ratio of tunable parameters and inference latency in Figure~\ref{fig:exp_efficiency}. Inference latency is calculated by averaging the inference time on the number of inference samples. MOELoRA(M) denotes the variant of MOELoRA, where every MOELoRA layer is accompanied by a task-motivated gate. The results show that MOELoRA achieves as high training and inference efficiency as LoRA (Full), which can save resources by training no more than $0.48$\% parameters of LLMs. MoLoRA and MOELoRA(M) need more trainable parameters, because they set the extra gates for each trainable low-rank layer. In terms of inference, all models need the same inference latency, except MoLoRA. The reason lies in that MoLoRA cannot recover the fine-tuned parameters as Equation~\eqref{eq5}, because the expert weights vary from samples. Therefore, it needs to accompany the MoLoRA layers when inference, which leads to more inference latency caused by the additional forward computation. This comparison indicates the benefit of the design of the task-motivated gate. As a response to the \textbf{RQ4}, the designed MOELoRA achieves high training and inference efficiency, and avoids efficiency degradation by the task-motivated gate.

\subsection{Case Study (RQ5)}
\begin{figure}[!t]
\centering
% \vspace{-1mm}
\includegraphics[width=1\linewidth]{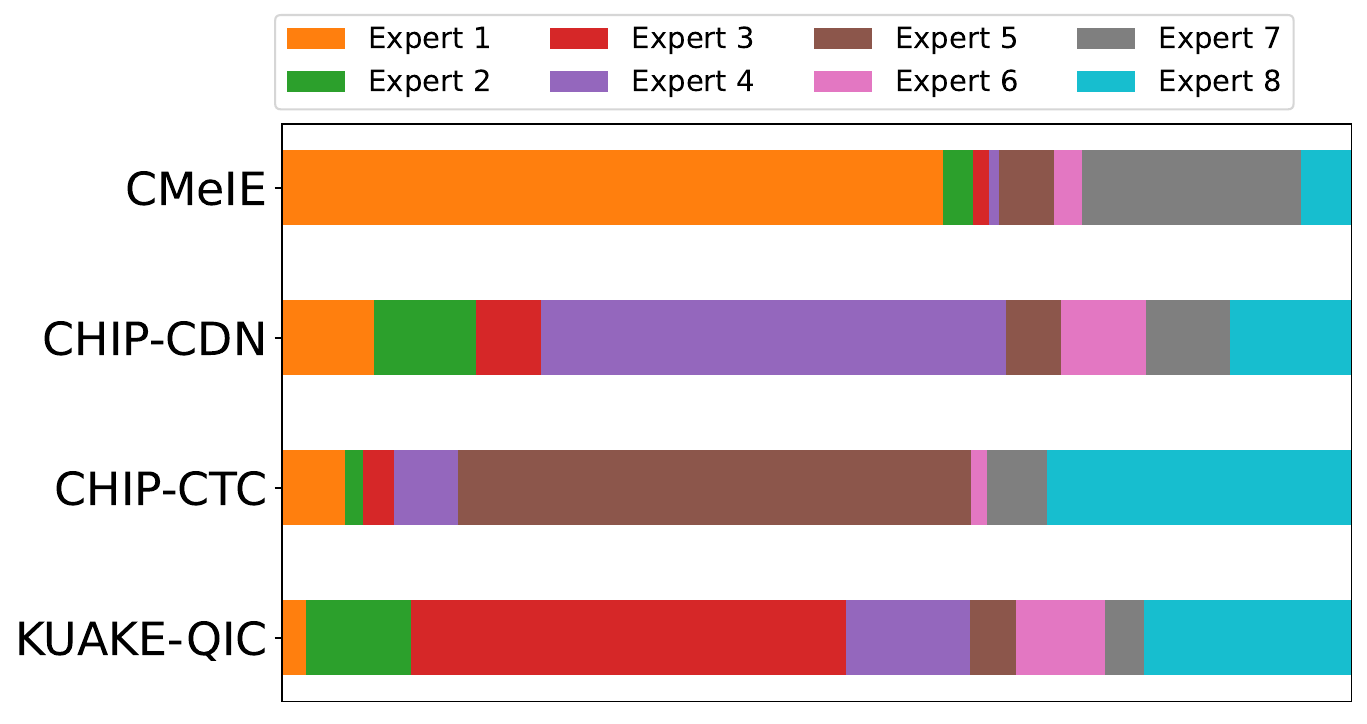}
\caption{The visualization of expert weights for various tasks. In each task, the length of the bar in different colors represents the weights for the corresponding expert.}
%\zys{需要更加详细的说明}}
\label{fig:exp_weight}
\vspace{-5mm}
\end{figure}

For \textbf{RQ4}, we present a visualization of the expert weights across four tasks in Figure~\ref{fig:exp_weight}. For each task, the length of the bar in different colors represents the weights for the corresponding expert. Since the expert weights are normalized to 1, the lengths of the bar for each task are the same. At a macro level, it is evident that the contributions from each expert vary significantly, underscoring the idea that different experts specialize in distinct facets of medical knowledge. Moreover, the pronounced disparities in weights across tasks highlight the diverse nature of medical applications. Taking a closer look at the tasks CHIP-CDN and KUAKE-QIC, we observe that their expert weights are largely congruent, with exceptions in experts 3 and 4. Considering diagnostic word normalization can bolster inquiry classification, the similarity in expert weights suggests that MOELoRA is adept at harnessing shared knowledge to benefit related tasks.
% Given that CHIP-CDN can be viewed as a precursor to KUAKE-QIC—since diagnostic word normalization can bolster inquiry classification—the similarity in expert weights suggests that MOELoRA is adept at harnessing shared knowledge to benefit intrinsically related tasks.

\section{Related Works}

\subsection{LLM for Medical Applications}
%\xdr{
Recently, the powerful capabilities of LLMs have been proven in many fields~\cite{fan2023recommender,xu2024multi,li2023agent4ranking,zheng2024harnessing,fu2023unified,luo2023recranker,luo2024integrating,li2023e4srec,wang2023plate,wang2023large}, including the medical domain. For instance, Med-PaLM~\cite{singhal2022large} proposes a new benchmark called MultiMedQA, which combines seven medical question-answering datasets to address the challenges of evaluating the clinical knowledge of LLM. 
Med-PaLM2 \cite{singhal2023towards} has further improved upon Med-PaLM by introducing a new prompting strategy called ensemble refinement. This strategy is based on CoT \cite{wei2022chain} and self-consistency \cite{wang2022self} and has shown significant improvements in MedQA. 
Then, ChatDoctor \cite{yunxiang2023chatdoctor} trains medical LLMs by 100,000 patient-doctor dialogues collected from a widely used online medical consultation platform. 
% The dataset is fine-tuned on LLaMA, and an automated information retrieval method is proposed to utilize online information, like Wikipedia.
Besides, HuaTuo \cite{wang2023huatuo} is initially based on LlaMA~\cite{touvron2023llama} and fine-tuned using Chinese medical knowledge from CMeKG \cite{byambasuren2019preliminary}. 
% Additionally, HuaTuo has introduced a new evaluation metric called SUS for Safety, Usability, and Smoothness. 
For more specific medical applications, Liu~\etal~\cite{liu2024large} proposes an LLM-based medication recommendation model, while Xu~\etal~\cite{xu2024editing} designs a model editing method to resolve hallucination in medical LLMs.
% Previous studies have demonstrated the potential of LLMs in the medical field. 
However, most previous works tend to focus on medical dialogue or one specific medical task while neglecting multiple important tasks simultaneously. Besides, they usually demand a significant fine-tuning cost for achieving generalization ability. 
%}

\subsection{Parameter Efficient Fine-tuning}
%\xdr{
Parameter efficient fine-tuning (PEFT) aims to improve the performance of LLMs on new tasks by minimizing the number of fine-tuning parameters and computational complexity. Adapter Tuning \cite{houlsby2019parameter} first introduces a lightweight adapter module, which has only a few trainable parameters.
% and has shown comparable results to fine-tuning on the top layers of LLMs. 
Prefix-tuning \cite{li-liang-2021-prefix} and P-Tuning \cite{liu-etal-2022-p} both construct a task-specific virtual token that adds trainable, continuous prompts or embeddings to the original text sequence.
%, making optimization more feasible than with discrete prompts. 
However, using prompts can be challenging for training and can also limit the available sequence length of the model. LoRA \cite{hu2021lora} introduces two trainable low-rank matrices into each dense layer. It has been shown to achieve comparable performance to full fine-tuning while requiring no additional computation during inference. Nevertheless, the LoRA fine-tuning performs inferiorly for multi-task medical applications. 
% It can only learn integral updated parameters for all tasks, which loses the vital task-specific information. 
In recent times, a thread of research named cross-task generalization~\cite{huang2023lorahub,sun2023multitask,asai2022attempt,ustun2022hyper} emerges, which proposes various parameter efficient fine-tuning strategies. However, different from the multi-task setting in this paper, they first train the model on too many tasks and aim to transfer the ability to unseen tasks. Due to the distinct setting, their method is difficult to be adapted to our problem. In a word, the multi-task parameter efficient fine-tuning for LLM-driven medical applications is still underexplored, and we take the first step.

%}

%no multi-task learning has been proposed. only a few cross-task generalizations work.

%\subsection{Multi-task Learning}

\section{Conclusion}

In this paper, we take the first step to explore the multi-task parameter efficient fine-tuning method for LLM-driven medical applications. To satisfy the requirements of efficiency for fine-tuning and effectiveness for multi-task, we propose a novel multi-task fine-tuning framework. Specifically, we design the MOELoRA architecture, which consists of several low-rank experts as the trainable parameters to learn task-related knowledge and retain high efficiency. Besides, a task-motivated gate function is devised to produce distinct fine-tuned parameters for various tasks. By the comprehensive experiments on a multi-task Chinese medical dataset, we verify the effectiveness of the proposed MOELoRA. In the future, we will further explore how to combine explicit medical knowledge, such as knowledge graphs, with LLMs by fine-tuning.

% \begin{acks}
% \end{acks}

\begin{acks}
    This research was supported by Tencent (CCF-Tencent Open Fund), Research Impact Fund (No.R1015-23), APRC - CityU New Research Initiatives (No.9610565, Start-up Grant for New Faculty of CityU), CityU - HKIDS Early Career Research Grant (No.9360163), Hong Kong ITC Innovation and Technology Fund Midstream Research Programme for Universities Project (No.ITS/034/22MS), Hong Kong Environmental and Conservation Fund (No. 88/2022), SIRG - CityU Strategic Interdisciplinary Research Grant (No.7020046, No.7020074).  %% Xiangyu's project
    % National Key Research and Development Program of China (2022ZD0117102), National Natural Science Foundation of China (No.62293551, No.62177038, No.62277042, No.62137002, No.61721002, No.61937001, No.62377038), Project of China Knowledge Centre for Engineering Science and Technology, ``LENOVO-XJTU'' Intelligent Industry Joint Laboratory Project. %% Feng's Project
\end{acks}

\bibliographystyle{ACM-Reference-Format}
\bibliography{main}


\begin{thebibliography}{63}

%%% the Web address.

\ifx \showCODEN    \undefined \def \showCODEN     #1{\unskip}     \fi
\ifx \showDOI      \undefined \def \showDOI       #1{#1}\fi
\ifx \showISBNx    \undefined \def \showISBNx     #1{\unskip}     \fi
\ifx \showISBNxiii \undefined \def \showISBNxiii  #1{\unskip}     \fi
\ifx \showISSN     \undefined \def \showISSN      #1{\unskip}     \fi
\ifx \showLCCN     \undefined \def \showLCCN      #1{\unskip}     \fi
\ifx \shownote     \undefined \def \shownote      #1{#1}          \fi
\ifx \showarticletitle \undefined \def \showarticletitle #1{#1}   \fi
\ifx \showURL      \undefined \def \showURL       {\relax}        \fi
% invisible to TeX
\providecommand\bibfield[2]{#2}
\providecommand\bibinfo[2]{#2}
\providecommand\natexlab[1]{#1}
\providecommand\showeprint[2][]{arXiv:#2}

\bibitem[Aghajanyan et~al\mbox{.}(2021)]%
        {aghajanyan2021intrinsic}
\bibfield{author}{\bibinfo{person}{Armen Aghajanyan}, \bibinfo{person}{Sonal Gupta}, {and} \bibinfo{person}{Luke Zettlemoyer}.} \bibinfo{year}{2021}\natexlab{}.
\newblock \showarticletitle{Intrinsic Dimensionality Explains the Effectiveness of Language Model Fine-Tuning}. In \bibinfo{booktitle}{\emph{Proceedings of the 59th Annual Meeting of the Association for Computational Linguistics and the 11th International Joint Conference on Natural Language Processing (Volume 1: Long Papers)}}. \bibinfo{pages}{7319--7328}.
\newblock


\bibitem[Asai et~al\mbox{.}(2022)]%
        {asai2022attempt}
\bibfield{author}{\bibinfo{person}{Akari Asai}, \bibinfo{person}{Mohammadreza Salehi}, \bibinfo{person}{Matthew~E Peters}, {and} \bibinfo{person}{Hannaneh Hajishirzi}.} \bibinfo{year}{2022}\natexlab{}.
\newblock \showarticletitle{Attempt: Parameter-efficient multi-task tuning via attentional mixtures of soft prompts}. In \bibinfo{booktitle}{\emph{Proceedings of the 2022 Conference on Empirical Methods in Natural Language Processing}}. \bibinfo{pages}{6655--6672}.
\newblock


\bibitem[Brown et~al\mbox{.}(2020)]%
        {brown2020language}
\bibfield{author}{\bibinfo{person}{Tom Brown}, \bibinfo{person}{Benjamin Mann}, \bibinfo{person}{Nick Ryder}, \bibinfo{person}{Melanie Subbiah}, \bibinfo{person}{Jared~D Kaplan}, \bibinfo{person}{Prafulla Dhariwal}, \bibinfo{person}{Arvind Neelakantan}, \bibinfo{person}{Pranav Shyam}, \bibinfo{person}{Girish Sastry}, \bibinfo{person}{Amanda Askell}, {et~al\mbox{.}}} \bibinfo{year}{2020}\natexlab{}.
\newblock \showarticletitle{Language models are few-shot learners}.
\newblock \bibinfo{journal}{\emph{Advances in neural information processing systems}}  \bibinfo{volume}{33} (\bibinfo{year}{2020}), \bibinfo{pages}{1877--1901}.
\newblock


\bibitem[Byambasuren et~al\mbox{.}(2019)]%
        {byambasuren2019preliminary}
\bibfield{author}{\bibinfo{person}{Odma Byambasuren}, \bibinfo{person}{Yunfei Yang}, \bibinfo{person}{Zhifang Sui}, \bibinfo{person}{Damai Dai}, \bibinfo{person}{Baobao Chang}, \bibinfo{person}{Sujian Li}, {and} \bibinfo{person}{Hongying Zan}.} \bibinfo{year}{2019}\natexlab{}.
\newblock \showarticletitle{Preliminary study on the construction of Chinese medical knowledge graph}.
\newblock \bibinfo{journal}{\emph{Journal of Chinese Information Processing}} \bibinfo{volume}{33}, \bibinfo{number}{10} (\bibinfo{year}{2019}), \bibinfo{pages}{1--9}.
\newblock


\bibitem[Crawshaw(2020)]%
        {crawshaw2020multi}
\bibfield{author}{\bibinfo{person}{Michael Crawshaw}.} \bibinfo{year}{2020}\natexlab{}.
\newblock \showarticletitle{Multi-task learning with deep neural networks: A survey}.
\newblock \bibinfo{journal}{\emph{arXiv preprint arXiv:2009.09796}} (\bibinfo{year}{2020}).
\newblock


\bibitem[Dong et~al\mbox{.}(2022)]%
        {dong2022survey}
\bibfield{author}{\bibinfo{person}{Qingxiu Dong}, \bibinfo{person}{Lei Li}, \bibinfo{person}{Damai Dai}, \bibinfo{person}{Ce Zheng}, \bibinfo{person}{Zhiyong Wu}, \bibinfo{person}{Baobao Chang}, \bibinfo{person}{Xu Sun}, \bibinfo{person}{Jingjing Xu}, {and} \bibinfo{person}{Zhifang Sui}.} \bibinfo{year}{2022}\natexlab{}.
\newblock \showarticletitle{A survey for in-context learning}.
\newblock \bibinfo{journal}{\emph{arXiv preprint arXiv:2301.00234}} (\bibinfo{year}{2022}).
\newblock


\bibitem[Du et~al\mbox{.}(2022)]%
        {du2022glm}
\bibfield{author}{\bibinfo{person}{Zhengxiao Du}, \bibinfo{person}{Yujie Qian}, \bibinfo{person}{Xiao Liu}, \bibinfo{person}{Ming Ding}, \bibinfo{person}{Jiezhong Qiu}, \bibinfo{person}{Zhilin Yang}, {and} \bibinfo{person}{Jie Tang}.} \bibinfo{year}{2022}\natexlab{}.
\newblock \showarticletitle{GLM: General Language Model Pretraining with Autoregressive Blank Infilling}. In \bibinfo{booktitle}{\emph{Proceedings of the 60th Annual Meeting of the Association for Computational Linguistics (Volume 1: Long Papers)}}. \bibinfo{pages}{320--335}.
\newblock


\bibitem[Fan et~al\mbox{.}(2022)]%
        {fan2022comprehensive}
\bibfield{author}{\bibinfo{person}{Wenqi Fan}, \bibinfo{person}{Xiangyu Zhao}, \bibinfo{person}{Xiao Chen}, \bibinfo{person}{Jingran Su}, \bibinfo{person}{Jingtong Gao}, \bibinfo{person}{Lin Wang}, \bibinfo{person}{Qidong Liu}, \bibinfo{person}{Yiqi Wang}, \bibinfo{person}{Han Xu}, \bibinfo{person}{Lei Chen}, {et~al\mbox{.}}} \bibinfo{year}{2022}\natexlab{}.
\newblock \showarticletitle{A comprehensive survey on trustworthy recommender systems}.
\newblock \bibinfo{journal}{\emph{arXiv preprint arXiv:2209.10117}} (\bibinfo{year}{2022}).
\newblock


\bibitem[Fan et~al\mbox{.}(2023)]%
        {fan2023recommender}
\bibfield{author}{\bibinfo{person}{Wenqi Fan}, \bibinfo{person}{Zihuai Zhao}, \bibinfo{person}{Jiatong Li}, \bibinfo{person}{Yunqing Liu}, \bibinfo{person}{Xiaowei Mei}, \bibinfo{person}{Yiqi Wang}, \bibinfo{person}{Jiliang Tang}, {and} \bibinfo{person}{Qing Li}.} \bibinfo{year}{2023}\natexlab{}.
\newblock \showarticletitle{Recommender systems in the era of large language models (llms)}.
\newblock \bibinfo{journal}{\emph{arXiv preprint arXiv:2307.02046}} (\bibinfo{year}{2023}).
\newblock


\bibitem[Fu et~al\mbox{.}(2023)]%
        {fu2023unified}
\bibfield{author}{\bibinfo{person}{Zichuan Fu}, \bibinfo{person}{Xiangyang Li}, \bibinfo{person}{Chuhan Wu}, \bibinfo{person}{Yichao Wang}, \bibinfo{person}{Kuicai Dong}, \bibinfo{person}{Xiangyu Zhao}, \bibinfo{person}{Mengchen Zhao}, \bibinfo{person}{Huifeng Guo}, {and} \bibinfo{person}{Ruiming Tang}.} \bibinfo{year}{2023}\natexlab{}.
\newblock \showarticletitle{A Unified Framework for Multi-Domain CTR Prediction via Large Language Models}.
\newblock \bibinfo{journal}{\emph{arXiv preprint arXiv:2312.10743}} (\bibinfo{year}{2023}).
\newblock


\bibitem[Gupta et~al\mbox{.}(2022)]%
        {gupta2022sparsely}
\bibfield{author}{\bibinfo{person}{Shashank Gupta}, \bibinfo{person}{Subhabrata Mukherjee}, \bibinfo{person}{Krishan Subudhi}, \bibinfo{person}{Eduardo Gonzalez}, \bibinfo{person}{Damien Jose}, \bibinfo{person}{Ahmed~H Awadallah}, {and} \bibinfo{person}{Jianfeng Gao}.} \bibinfo{year}{2022}\natexlab{}.
\newblock \showarticletitle{Sparsely activated mixture-of-experts are robust multi-task learners}.
\newblock \bibinfo{journal}{\emph{arXiv preprint arXiv:2204.07689}} (\bibinfo{year}{2022}).
\newblock


\bibitem[Hadi et~al\mbox{.}(2023)]%
        {hadi2023survey}
\bibfield{author}{\bibinfo{person}{Muhammad~Usman Hadi}, \bibinfo{person}{R Qureshi}, \bibinfo{person}{A Shah}, \bibinfo{person}{M Irfan}, \bibinfo{person}{A Zafar}, \bibinfo{person}{MB Shaikh}, \bibinfo{person}{N Akhtar}, \bibinfo{person}{J Wu}, {and} \bibinfo{person}{S Mirjalili}.} \bibinfo{year}{2023}\natexlab{}.
\newblock \showarticletitle{A survey on large language models: Applications, challenges, limitations, and practical usage}.
\newblock \bibinfo{journal}{\emph{TechRxiv}} (\bibinfo{year}{2023}).
\newblock


\bibitem[Houlsby et~al\mbox{.}(2019)]%
        {houlsby2019parameter}
\bibfield{author}{\bibinfo{person}{Neil Houlsby}, \bibinfo{person}{Andrei Giurgiu}, \bibinfo{person}{Stanislaw Jastrzebski}, \bibinfo{person}{Bruna Morrone}, \bibinfo{person}{Quentin De~Laroussilhe}, \bibinfo{person}{Andrea Gesmundo}, \bibinfo{person}{Mona Attariyan}, {and} \bibinfo{person}{Sylvain Gelly}.} \bibinfo{year}{2019}\natexlab{}.
\newblock \showarticletitle{Parameter-efficient transfer learning for NLP}. In \bibinfo{booktitle}{\emph{International Conference on Machine Learning}}. PMLR, \bibinfo{pages}{2790--2799}.
\newblock


\bibitem[Hu et~al\mbox{.}(2021)]%
        {hu2021lora}
\bibfield{author}{\bibinfo{person}{Edward~J Hu}, \bibinfo{person}{Phillip Wallis}, \bibinfo{person}{Zeyuan Allen-Zhu}, \bibinfo{person}{Yuanzhi Li}, \bibinfo{person}{Shean Wang}, \bibinfo{person}{Lu Wang}, \bibinfo{person}{Weizhu Chen}, {et~al\mbox{.}}} \bibinfo{year}{2021}\natexlab{}.
\newblock \showarticletitle{LoRA: Low-Rank Adaptation of Large Language Models}. In \bibinfo{booktitle}{\emph{International Conference on Learning Representations}}.
\newblock


\bibitem[Huang et~al\mbox{.}(2023)]%
        {huang2023lorahub}
\bibfield{author}{\bibinfo{person}{Chengsong Huang}, \bibinfo{person}{Qian Liu}, \bibinfo{person}{Bill~Yuchen Lin}, \bibinfo{person}{Tianyu Pang}, \bibinfo{person}{Chao Du}, {and} \bibinfo{person}{Min Lin}.} \bibinfo{year}{2023}\natexlab{}.
\newblock \showarticletitle{LoraHub: Efficient Cross-Task Generalization via Dynamic LoRA Composition}.
\newblock \bibinfo{journal}{\emph{arXiv preprint arXiv:2307.13269}} (\bibinfo{year}{2023}).
\newblock


\bibitem[Ilharco et~al\mbox{.}(2022)]%
        {ilharco2022editing}
\bibfield{author}{\bibinfo{person}{Gabriel Ilharco}, \bibinfo{person}{Marco~Tulio Ribeiro}, \bibinfo{person}{Mitchell Wortsman}, \bibinfo{person}{Ludwig Schmidt}, \bibinfo{person}{Hannaneh Hajishirzi}, {and} \bibinfo{person}{Ali Farhadi}.} \bibinfo{year}{2022}\natexlab{}.
\newblock \showarticletitle{Editing models with task arithmetic}. In \bibinfo{booktitle}{\emph{The Eleventh International Conference on Learning Representations}}.
\newblock


\bibitem[Kenton and Toutanova(2019)]%
        {kenton2019bert}
\bibfield{author}{\bibinfo{person}{Jacob Devlin Ming-Wei~Chang Kenton} {and} \bibinfo{person}{Lee~Kristina Toutanova}.} \bibinfo{year}{2019}\natexlab{}.
\newblock \showarticletitle{BERT: Pre-training of Deep Bidirectional Transformers for Language Understanding}. In \bibinfo{booktitle}{\emph{Proceedings of NAACL-HLT}}. \bibinfo{pages}{4171--4186}.
\newblock


\bibitem[Li et~al\mbox{.}(2023a)]%
        {li2023e4srec}
\bibfield{author}{\bibinfo{person}{Xinhang Li}, \bibinfo{person}{Chong Chen}, \bibinfo{person}{Xiangyu Zhao}, \bibinfo{person}{Yong Zhang}, {and} \bibinfo{person}{Chunxiao Xing}.} \bibinfo{year}{2023}\natexlab{a}.
\newblock \showarticletitle{E4SRec: An elegant effective efficient extensible solution of large language models for sequential recommendation}.
\newblock \bibinfo{journal}{\emph{arXiv preprint arXiv:2312.02443}} (\bibinfo{year}{2023}).
\newblock


\bibitem[Li et~al\mbox{.}(2023b)]%
        {li2023agent4ranking}
\bibfield{author}{\bibinfo{person}{Xiaopeng Li}, \bibinfo{person}{Lixin Su}, \bibinfo{person}{Pengyue Jia}, \bibinfo{person}{Xiangyu Zhao}, \bibinfo{person}{Suqi Cheng}, \bibinfo{person}{Junfeng Wang}, {and} \bibinfo{person}{Dawei Yin}.} \bibinfo{year}{2023}\natexlab{b}.
\newblock \showarticletitle{Agent4Ranking: Semantic Robust Ranking via Personalized Query Rewriting Using Multi-agent LLM}.
\newblock \bibinfo{journal}{\emph{arXiv preprint arXiv:2312.15450}} (\bibinfo{year}{2023}).
\newblock


\bibitem[Li et~al\mbox{.}(2020)]%
        {li2020dice}
\bibfield{author}{\bibinfo{person}{Xiaoya Li}, \bibinfo{person}{Xiaofei Sun}, \bibinfo{person}{Yuxian Meng}, \bibinfo{person}{Junjun Liang}, \bibinfo{person}{Fei Wu}, {and} \bibinfo{person}{Jiwei Li}.} \bibinfo{year}{2020}\natexlab{}.
\newblock \showarticletitle{Dice Loss for Data-imbalanced NLP Tasks}. In \bibinfo{booktitle}{\emph{Proceedings of the 58th Annual Meeting of the Association for Computational Linguistics}}. \bibinfo{pages}{465--476}.
\newblock


\bibitem[Li and Liang(2021)]%
        {li-liang-2021-prefix}
\bibfield{author}{\bibinfo{person}{Xiang~Lisa Li} {and} \bibinfo{person}{Percy Liang}.} \bibinfo{year}{2021}\natexlab{}.
\newblock \showarticletitle{Prefix-Tuning: Optimizing Continuous Prompts for Generation}. In \bibinfo{booktitle}{\emph{Proceedings of the 59th Annual Meeting of the Association for Computational Linguistics and the 11th International Joint Conference on Natural Language Processing (Volume 1: Long Papers)}}. \bibinfo{publisher}{Association for Computational Linguistics}, \bibinfo{address}{Online}, \bibinfo{pages}{4582--4597}.
\newblock
\urldef\tempurl%
\url{https://doi.org/10.18653/v1/2021.acl-long.353}
\showDOI{\tempurl}


\bibitem[Lin and Och(2004)]%
        {lin2004automatic}
\bibfield{author}{\bibinfo{person}{Chin-Yew Lin} {and} \bibinfo{person}{Franz~Josef Och}.} \bibinfo{year}{2004}\natexlab{}.
\newblock \showarticletitle{Automatic evaluation of machine translation quality using longest common subsequence and skip-bigram statistics}. In \bibinfo{booktitle}{\emph{Proceedings of the 42nd Annual Meeting of the Association for Computational Linguistics (ACL-04)}}. \bibinfo{pages}{605--612}.
\newblock


\bibitem[Liu et~al\mbox{.}(2024)]%
        {liu2024large}
\bibfield{author}{\bibinfo{person}{Qidong Liu}, \bibinfo{person}{Xian Wu}, \bibinfo{person}{Xiangyu Zhao}, \bibinfo{person}{Yuanshao Zhu}, \bibinfo{person}{Zijian Zhang}, \bibinfo{person}{Feng Tian}, {and} \bibinfo{person}{Yefeng Zheng}.} \bibinfo{year}{2024}\natexlab{}.
\newblock \showarticletitle{Large Language Model Distilling Medication Recommendation Model}.
\newblock \bibinfo{journal}{\emph{arXiv preprint arXiv:2402.02803}} (\bibinfo{year}{2024}).
\newblock


\bibitem[Liu et~al\mbox{.}(2022)]%
        {liu-etal-2022-p}
\bibfield{author}{\bibinfo{person}{Xiao Liu}, \bibinfo{person}{Kaixuan Ji}, \bibinfo{person}{Yicheng Fu}, \bibinfo{person}{Weng Tam}, \bibinfo{person}{Zhengxiao Du}, \bibinfo{person}{Zhilin Yang}, {and} \bibinfo{person}{Jie Tang}.} \bibinfo{year}{2022}\natexlab{}.
\newblock \showarticletitle{{P}-Tuning: Prompt Tuning Can Be Comparable to Fine-tuning Across Scales and Tasks}. In \bibinfo{booktitle}{\emph{Proceedings of the 60th Annual Meeting of the Association for Computational Linguistics (Volume 2: Short Papers)}}. \bibinfo{publisher}{Association for Computational Linguistics}, \bibinfo{address}{Dublin, Ireland}, \bibinfo{pages}{61--68}.
\newblock
\urldef\tempurl%
\url{https://doi.org/10.18653/v1/2022.acl-short.8}
\showDOI{\tempurl}


\bibitem[Liu et~al\mbox{.}(2023b)]%
        {liu2023gpt}
\bibfield{author}{\bibinfo{person}{Xiao Liu}, \bibinfo{person}{Yanan Zheng}, \bibinfo{person}{Zhengxiao Du}, \bibinfo{person}{Ming Ding}, \bibinfo{person}{Yujie Qian}, \bibinfo{person}{Zhilin Yang}, {and} \bibinfo{person}{Jie Tang}.} \bibinfo{year}{2023}\natexlab{b}.
\newblock \showarticletitle{GPT understands, too}.
\newblock \bibinfo{journal}{\emph{AI Open}} (\bibinfo{year}{2023}).
\newblock


\bibitem[Liu et~al\mbox{.}(2023a)]%
        {liu2023multi}
\bibfield{author}{\bibinfo{person}{Ziru Liu}, \bibinfo{person}{Jiejie Tian}, \bibinfo{person}{Qingpeng Cai}, \bibinfo{person}{Xiangyu Zhao}, \bibinfo{person}{Jingtong Gao}, \bibinfo{person}{Shuchang Liu}, \bibinfo{person}{Dayou Chen}, \bibinfo{person}{Tonghao He}, \bibinfo{person}{Dong Zheng}, \bibinfo{person}{Peng Jiang}, {et~al\mbox{.}}} \bibinfo{year}{2023}\natexlab{a}.
\newblock \showarticletitle{Multi-task recommendations with reinforcement learning}. In \bibinfo{booktitle}{\emph{Proceedings of the ACM Web Conference 2023}}. \bibinfo{pages}{1273--1282}.
\newblock


\bibitem[Luo et~al\mbox{.}(2023)]%
        {luo2023recranker}
\bibfield{author}{\bibinfo{person}{Sichun Luo}, \bibinfo{person}{Bowei He}, \bibinfo{person}{Haohan Zhao}, \bibinfo{person}{Yinya Huang}, \bibinfo{person}{Aojun Zhou}, \bibinfo{person}{Zongpeng Li}, \bibinfo{person}{Yuanzhang Xiao}, \bibinfo{person}{Mingjie Zhan}, {and} \bibinfo{person}{Linqi Song}.} \bibinfo{year}{2023}\natexlab{}.
\newblock \showarticletitle{RecRanker: Instruction Tuning Large Language Model as Ranker for Top-k Recommendation}.
\newblock \bibinfo{journal}{\emph{arXiv preprint arXiv:2312.16018}} (\bibinfo{year}{2023}).
\newblock


\bibitem[Luo et~al\mbox{.}(2024)]%
        {luo2024integrating}
\bibfield{author}{\bibinfo{person}{Sichun Luo}, \bibinfo{person}{Yuxuan Yao}, \bibinfo{person}{Bowei He}, \bibinfo{person}{Yinya Huang}, \bibinfo{person}{Aojun Zhou}, \bibinfo{person}{Xinyi Zhang}, \bibinfo{person}{Yuanzhang Xiao}, \bibinfo{person}{Mingjie Zhan}, {and} \bibinfo{person}{Linqi Song}.} \bibinfo{year}{2024}\natexlab{}.
\newblock \showarticletitle{Integrating Large Language Models into Recommendation via Mutual Augmentation and Adaptive Aggregation}.
\newblock \bibinfo{journal}{\emph{arXiv preprint arXiv:2401.13870}} (\bibinfo{year}{2024}).
\newblock


\bibitem[Miura et~al\mbox{.}(2020)]%
        {miura2020improving}
\bibfield{author}{\bibinfo{person}{Yasuhide Miura}, \bibinfo{person}{Yuhao Zhang}, \bibinfo{person}{Emily~Bao Tsai}, \bibinfo{person}{Curtis~P Langlotz}, {and} \bibinfo{person}{Dan Jurafsky}.} \bibinfo{year}{2020}\natexlab{}.
\newblock \showarticletitle{Improving factual completeness and consistency of image-to-text radiology report generation}.
\newblock \bibinfo{journal}{\emph{arXiv preprint arXiv:2010.10042}} (\bibinfo{year}{2020}).
\newblock


\bibitem[OpenAI(2023)]%
        {openai2023gpt4}
\bibfield{author}{\bibinfo{person}{OpenAI}.} \bibinfo{year}{2023}\natexlab{}.
\newblock \showarticletitle{GPT-4 Technical Report}.
\newblock \bibinfo{journal}{\emph{arXiv preprint arXiv:2303.08774}} (\bibinfo{year}{2023}).
\newblock


\bibitem[Opitz and Burst(2019)]%
        {opitz2019macro}
\bibfield{author}{\bibinfo{person}{Juri Opitz} {and} \bibinfo{person}{Sebastian Burst}.} \bibinfo{year}{2019}\natexlab{}.
\newblock \showarticletitle{Macro f1 and macro f1}.
\newblock \bibinfo{journal}{\emph{arXiv preprint arXiv:1911.03347}} (\bibinfo{year}{2019}).
\newblock


\bibitem[Qiao et~al\mbox{.}(2019)]%
        {Qiao2019MNNMA}
\bibfield{author}{\bibinfo{person}{Zhi Qiao}, \bibinfo{person}{X. Wu}, \bibinfo{person}{Shen Ge}, {and} \bibinfo{person}{Wei Fan}.} \bibinfo{year}{2019}\natexlab{}.
\newblock \showarticletitle{MNN: Multimodal Attentional Neural Networks for Diagnosis Prediction}. In \bibinfo{booktitle}{\emph{International Joint Conference on Artificial Intelligence}}.
\newblock
\urldef\tempurl%
\url{https://api.semanticscholar.org/CorpusID:199466261}
\showURL{%
\tempurl}


\bibitem[Rezayi et~al\mbox{.}(2022)]%
        {rezayi2022clinicalradiobert}
\bibfield{author}{\bibinfo{person}{Saed Rezayi}, \bibinfo{person}{Haixing Dai}, \bibinfo{person}{Zhengliang Liu}, \bibinfo{person}{Zihao Wu}, \bibinfo{person}{Akarsh Hebbar}, \bibinfo{person}{Andrew~H Burns}, \bibinfo{person}{Lin Zhao}, \bibinfo{person}{Dajiang Zhu}, \bibinfo{person}{Quanzheng Li}, \bibinfo{person}{Wei Liu}, {et~al\mbox{.}}} \bibinfo{year}{2022}\natexlab{}.
\newblock \showarticletitle{Clinicalradiobert: Knowledge-infused few shot learning for clinical notes named entity recognition}. In \bibinfo{booktitle}{\emph{International Workshop on Machine Learning in Medical Imaging}}. Springer, \bibinfo{pages}{269--278}.
\newblock


\bibitem[Riquelme et~al\mbox{.}(2021)]%
        {riquelme2021scaling}
\bibfield{author}{\bibinfo{person}{Carlos Riquelme}, \bibinfo{person}{Joan Puigcerver}, \bibinfo{person}{Basil Mustafa}, \bibinfo{person}{Maxim Neumann}, \bibinfo{person}{Rodolphe Jenatton}, \bibinfo{person}{Andr{\'e} Susano~Pinto}, \bibinfo{person}{Daniel Keysers}, {and} \bibinfo{person}{Neil Houlsby}.} \bibinfo{year}{2021}\natexlab{}.
\newblock \showarticletitle{Scaling vision with sparse mixture of experts}.
\newblock \bibinfo{journal}{\emph{Advances in Neural Information Processing Systems}}  \bibinfo{volume}{34} (\bibinfo{year}{2021}), \bibinfo{pages}{8583--8595}.
\newblock


\bibitem[Shazeer et~al\mbox{.}(2017)]%
        {shazeer2017outrageously}
\bibfield{author}{\bibinfo{person}{Noam Shazeer}, \bibinfo{person}{Azalia Mirhoseini}, \bibinfo{person}{Krzysztof Maziarz}, \bibinfo{person}{Andy Davis}, \bibinfo{person}{Quoc Le}, \bibinfo{person}{Geoffrey Hinton}, {and} \bibinfo{person}{Jeff Dean}.} \bibinfo{year}{2017}\natexlab{}.
\newblock \showarticletitle{Outrageously large neural networks: The sparsely-gated mixture-of-experts layer}.
\newblock \bibinfo{journal}{\emph{arXiv preprint arXiv:1701.06538}} (\bibinfo{year}{2017}).
\newblock


\bibitem[Sheng et~al\mbox{.}(2021)]%
        {sheng2021one}
\bibfield{author}{\bibinfo{person}{Xiang-Rong Sheng}, \bibinfo{person}{Liqin Zhao}, \bibinfo{person}{Guorui Zhou}, \bibinfo{person}{Xinyao Ding}, \bibinfo{person}{Binding Dai}, \bibinfo{person}{Qiang Luo}, \bibinfo{person}{Siran Yang}, \bibinfo{person}{Jingshan Lv}, \bibinfo{person}{Chi Zhang}, \bibinfo{person}{Hongbo Deng}, {et~al\mbox{.}}} \bibinfo{year}{2021}\natexlab{}.
\newblock \showarticletitle{One model to serve all: Star topology adaptive recommender for multi-domain ctr prediction}. In \bibinfo{booktitle}{\emph{Proceedings of the 30th ACM International Conference on Information \& Knowledge Management}}. \bibinfo{pages}{4104--4113}.
\newblock


\bibitem[Singhal et~al\mbox{.}(2023a)]%
        {singhal2022large}
\bibfield{author}{\bibinfo{person}{Karan Singhal}, \bibinfo{person}{Shekoofeh Azizi}, \bibinfo{person}{Tao Tu}, \bibinfo{person}{S~Sara Mahdavi}, \bibinfo{person}{Jason Wei}, \bibinfo{person}{Hyung~Won Chung}, \bibinfo{person}{Nathan Scales}, \bibinfo{person}{Ajay Tanwani}, \bibinfo{person}{Heather Cole-Lewis}, \bibinfo{person}{Stephen Pfohl}, {et~al\mbox{.}}} \bibinfo{year}{2023}\natexlab{a}.
\newblock \showarticletitle{Large language models encode clinical knowledge}.
\newblock \bibinfo{journal}{\emph{Nature}} \bibinfo{volume}{620}, \bibinfo{number}{7972} (\bibinfo{year}{2023}), \bibinfo{pages}{172--180}.
\newblock


\bibitem[Singhal et~al\mbox{.}(2023b)]%
        {singhal2023towards}
\bibfield{author}{\bibinfo{person}{Karan Singhal}, \bibinfo{person}{Tao Tu}, \bibinfo{person}{Juraj Gottweis}, \bibinfo{person}{Rory Sayres}, \bibinfo{person}{Ellery Wulczyn}, \bibinfo{person}{Le Hou}, \bibinfo{person}{Kevin Clark}, \bibinfo{person}{Stephen Pfohl}, \bibinfo{person}{Heather Cole-Lewis}, \bibinfo{person}{Darlene Neal}, {et~al\mbox{.}}} \bibinfo{year}{2023}\natexlab{b}.
\newblock \showarticletitle{Towards expert-level medical question answering with large language models}.
\newblock \bibinfo{journal}{\emph{arXiv preprint arXiv:2305.09617}} (\bibinfo{year}{2023}).
\newblock


\bibitem[Sun et~al\mbox{.}(2023)]%
        {sun2023multitask}
\bibfield{author}{\bibinfo{person}{Tianxiang Sun}, \bibinfo{person}{Zhengfu He}, \bibinfo{person}{Qin Zhu}, \bibinfo{person}{Xipeng Qiu}, {and} \bibinfo{person}{Xuan-Jing Huang}.} \bibinfo{year}{2023}\natexlab{}.
\newblock \showarticletitle{Multitask Pre-training of Modular Prompt for Chinese Few-Shot Learning}. In \bibinfo{booktitle}{\emph{Proceedings of the 61st Annual Meeting of the Association for Computational Linguistics (Volume 1: Long Papers)}}. \bibinfo{pages}{11156--11172}.
\newblock


\bibitem[Touvron et~al\mbox{.}(2023)]%
        {touvron2023llama}
\bibfield{author}{\bibinfo{person}{Hugo Touvron}, \bibinfo{person}{Thibaut Lavril}, \bibinfo{person}{Gautier Izacard}, \bibinfo{person}{Xavier Martinet}, \bibinfo{person}{Marie-Anne Lachaux}, \bibinfo{person}{Timoth{\'e}e Lacroix}, \bibinfo{person}{Baptiste Rozi{\`e}re}, \bibinfo{person}{Naman Goyal}, \bibinfo{person}{Eric Hambro}, \bibinfo{person}{Faisal Azhar}, \bibinfo{person}{Aurelien Rodriguez}, \bibinfo{person}{Armand Joulin}, \bibinfo{person}{Edouard Grave}, {and} \bibinfo{person}{Guillaume Lample}.} \bibinfo{year}{2023}\natexlab{}.
\newblock \showarticletitle{LLaMA: Open and Efficient Foundation Language Models}.
\newblock \bibinfo{journal}{\emph{arXiv preprint arXiv:2302.13971}} (\bibinfo{year}{2023}).
\newblock


\bibitem[{\"U}st{\"u}n et~al\mbox{.}(2022)]%
        {ustun2022hyper}
\bibfield{author}{\bibinfo{person}{Ahmet {\"U}st{\"u}n}, \bibinfo{person}{Arianna Bisazza}, \bibinfo{person}{Gosse Bouma}, \bibinfo{person}{Gertjan van Noord}, {and} \bibinfo{person}{Sebastian Ruder}.} \bibinfo{year}{2022}\natexlab{}.
\newblock \showarticletitle{Hyper-X: A unified hypernetwork for multi-task multilingual transfer}.
\newblock \bibinfo{journal}{\emph{arXiv preprint arXiv:2205.12148}} (\bibinfo{year}{2022}).
\newblock


\bibitem[Wang et~al\mbox{.}(2023b)]%
        {wang2023huatuo}
\bibfield{author}{\bibinfo{person}{Haochun Wang}, \bibinfo{person}{Chi Liu}, \bibinfo{person}{Nuwa Xi}, \bibinfo{person}{Zewen Qiang}, \bibinfo{person}{Sendong Zhao}, \bibinfo{person}{Bing Qin}, {and} \bibinfo{person}{Ting Liu}.} \bibinfo{year}{2023}\natexlab{b}.
\newblock \bibinfo{title}{HuaTuo: Tuning LLaMA Model with Chinese Medical Knowledge}.
\newblock
\newblock
\showeprint[arxiv]{2304.06975}~[cs.CL]


\bibitem[Wang et~al\mbox{.}(2023c)]%
        {wang2023survey}
\bibfield{author}{\bibinfo{person}{Lei Wang}, \bibinfo{person}{Chen Ma}, \bibinfo{person}{Xueyang Feng}, \bibinfo{person}{Zeyu Zhang}, \bibinfo{person}{Hao Yang}, \bibinfo{person}{Jingsen Zhang}, \bibinfo{person}{Zhiyuan Chen}, \bibinfo{person}{Jiakai Tang}, \bibinfo{person}{Xu Chen}, \bibinfo{person}{Yankai Lin}, {et~al\mbox{.}}} \bibinfo{year}{2023}\natexlab{c}.
\newblock \showarticletitle{A survey on large language model based autonomous agents}.
\newblock \bibinfo{journal}{\emph{arXiv preprint arXiv:2308.11432}} (\bibinfo{year}{2023}).
\newblock


\bibitem[Wang et~al\mbox{.}(2023e)]%
        {wang2023large}
\bibfield{author}{\bibinfo{person}{Maolin Wang}, \bibinfo{person}{Yao Zhao}, \bibinfo{person}{Jiajia Liu}, \bibinfo{person}{Jingdong Chen}, \bibinfo{person}{Chenyi Zhuang}, \bibinfo{person}{Jinjie Gu}, \bibinfo{person}{Ruocheng Guo}, {and} \bibinfo{person}{Xiangyu Zhao}.} \bibinfo{year}{2023}\natexlab{e}.
\newblock \showarticletitle{Large Multimodal Model Compression via Efficient Pruning and Distillation at AntGroup}.
\newblock \bibinfo{journal}{\emph{arXiv preprint arXiv:2312.05795}} (\bibinfo{year}{2023}).
\newblock


\bibitem[Wang et~al\mbox{.}(2022b)]%
        {wang2022self}
\bibfield{author}{\bibinfo{person}{Xuezhi Wang}, \bibinfo{person}{Jason Wei}, \bibinfo{person}{Dale Schuurmans}, \bibinfo{person}{Quoc Le}, \bibinfo{person}{Ed Chi}, \bibinfo{person}{Sharan Narang}, \bibinfo{person}{Aakanksha Chowdhery}, {and} \bibinfo{person}{Denny Zhou}.} \bibinfo{year}{2022}\natexlab{b}.
\newblock \showarticletitle{Self-consistency improves chain of thought reasoning in language models}.
\newblock \bibinfo{journal}{\emph{arXiv preprint arXiv:2203.11171}} (\bibinfo{year}{2022}).
\newblock


\bibitem[Wang et~al\mbox{.}(2023a)]%
        {wang2023multi}
\bibfield{author}{\bibinfo{person}{Yuhao Wang}, \bibinfo{person}{Ha~Tsz Lam}, \bibinfo{person}{Yi Wong}, \bibinfo{person}{Ziru Liu}, \bibinfo{person}{Xiangyu Zhao}, \bibinfo{person}{Yichao Wang}, \bibinfo{person}{Bo Chen}, \bibinfo{person}{Huifeng Guo}, {and} \bibinfo{person}{Ruiming Tang}.} \bibinfo{year}{2023}\natexlab{a}.
\newblock \showarticletitle{Multi-task deep recommender systems: A survey}.
\newblock \bibinfo{journal}{\emph{arXiv preprint arXiv:2302.03525}} (\bibinfo{year}{2023}).
\newblock


\bibitem[Wang et~al\mbox{.}(2022a)]%
        {wang2022nested}
\bibfield{author}{\bibinfo{person}{Yu Wang}, \bibinfo{person}{Hanghang Tong}, \bibinfo{person}{Ziye Zhu}, {and} \bibinfo{person}{Yun Li}.} \bibinfo{year}{2022}\natexlab{a}.
\newblock \showarticletitle{Nested named entity recognition: a survey}.
\newblock \bibinfo{journal}{\emph{ACM Transactions on Knowledge Discovery from Data (TKDD)}} \bibinfo{volume}{16}, \bibinfo{number}{6} (\bibinfo{year}{2022}), \bibinfo{pages}{1--29}.
\newblock


\bibitem[Wang et~al\mbox{.}(2023d)]%
        {wang2023plate}
\bibfield{author}{\bibinfo{person}{Yuhao Wang}, \bibinfo{person}{Xiangyu Zhao}, \bibinfo{person}{Bo Chen}, \bibinfo{person}{Qidong Liu}, \bibinfo{person}{Huifeng Guo}, \bibinfo{person}{Huanshuo Liu}, \bibinfo{person}{Yichao Wang}, \bibinfo{person}{Rui Zhang}, {and} \bibinfo{person}{Ruiming Tang}.} \bibinfo{year}{2023}\natexlab{d}.
\newblock \showarticletitle{PLATE: A Prompt-Enhanced Paradigm for Multi-Scenario Recommendations}. In \bibinfo{booktitle}{\emph{Proceedings of the 46th International ACM SIGIR Conference on Research and Development in Information Retrieval}}. \bibinfo{pages}{1498--1507}.
\newblock


\bibitem[Wei et~al\mbox{.}(2022)]%
        {wei2022chain}
\bibfield{author}{\bibinfo{person}{Jason Wei}, \bibinfo{person}{Xuezhi Wang}, \bibinfo{person}{Dale Schuurmans}, \bibinfo{person}{Maarten Bosma}, \bibinfo{person}{Fei Xia}, \bibinfo{person}{Ed Chi}, \bibinfo{person}{Quoc~V Le}, \bibinfo{person}{Denny Zhou}, {et~al\mbox{.}}} \bibinfo{year}{2022}\natexlab{}.
\newblock \showarticletitle{Chain-of-thought prompting elicits reasoning in large language models}.
\newblock \bibinfo{journal}{\emph{Advances in Neural Information Processing Systems}}  \bibinfo{volume}{35} (\bibinfo{year}{2022}), \bibinfo{pages}{24824--24837}.
\newblock


\bibitem[Xu et~al\mbox{.}(2024a)]%
        {xu2024multi}
\bibfield{author}{\bibinfo{person}{Derong Xu}, \bibinfo{person}{Ziheng Zhang}, \bibinfo{person}{Zhenxi Lin}, \bibinfo{person}{Xian Wu}, \bibinfo{person}{Zhihong Zhu}, \bibinfo{person}{Tong Xu}, \bibinfo{person}{Xiangyu Zhao}, \bibinfo{person}{Yefeng Zheng}, {and} \bibinfo{person}{Enhong Chen}.} \bibinfo{year}{2024}\natexlab{a}.
\newblock \showarticletitle{Multi-perspective Improvement of Knowledge Graph Completion with Large Language Models}.
\newblock \bibinfo{journal}{\emph{arXiv preprint arXiv:2403.01972}} (\bibinfo{year}{2024}).
\newblock


\bibitem[Xu et~al\mbox{.}(2024b)]%
        {xu2024editing}
\bibfield{author}{\bibinfo{person}{Derong Xu}, \bibinfo{person}{Ziheng Zhang}, \bibinfo{person}{Zhihong Zhu}, \bibinfo{person}{Zhenxi Lin}, \bibinfo{person}{Qidong Liu}, \bibinfo{person}{Xian Wu}, \bibinfo{person}{Tong Xu}, \bibinfo{person}{Xiangyu Zhao}, \bibinfo{person}{Yefeng Zheng}, {and} \bibinfo{person}{Enhong Chen}.} \bibinfo{year}{2024}\natexlab{b}.
\newblock \showarticletitle{Editing Factual Knowledge and Explanatory Ability of Medical Large Language Models}.
\newblock \bibinfo{journal}{\emph{arXiv preprint arXiv:2402.18099}} (\bibinfo{year}{2024}).
\newblock


\bibitem[Yang et~al\mbox{.}(2023)]%
        {yang2023baichuan}
\bibfield{author}{\bibinfo{person}{Aiyuan Yang}, \bibinfo{person}{Bin Xiao}, \bibinfo{person}{Bingning Wang}, \bibinfo{person}{Borong Zhang}, \bibinfo{person}{Chao Yin}, \bibinfo{person}{Chenxu Lv}, \bibinfo{person}{Da Pan}, \bibinfo{person}{Dian Wang}, \bibinfo{person}{Dong Yan}, \bibinfo{person}{Fan Yang}, {et~al\mbox{.}}} \bibinfo{year}{2023}\natexlab{}.
\newblock \showarticletitle{Baichuan 2: Open Large-scale Language Models}.
\newblock \bibinfo{journal}{\emph{arXiv preprint arXiv:2309.10305}} (\bibinfo{year}{2023}).
\newblock


\bibitem[Yunxiang et~al\mbox{.}(2023)]%
        {yunxiang2023chatdoctor}
\bibfield{author}{\bibinfo{person}{Li Yunxiang}, \bibinfo{person}{Li Zihan}, \bibinfo{person}{Zhang Kai}, \bibinfo{person}{Dan Ruilong}, {and} \bibinfo{person}{Zhang You}.} \bibinfo{year}{2023}\natexlab{}.
\newblock \showarticletitle{Chatdoctor: A medical chat model fine-tuned on llama model using medical domain knowledge}.
\newblock \bibinfo{journal}{\emph{arXiv preprint arXiv:2303.14070}} (\bibinfo{year}{2023}).
\newblock


\bibitem[Zadouri et~al\mbox{.}(2023)]%
        {zadouri2023pushing}
\bibfield{author}{\bibinfo{person}{Ted Zadouri}, \bibinfo{person}{Ahmet {\"U}st{\"u}n}, \bibinfo{person}{Arash Ahmadian}, \bibinfo{person}{Beyza Ermi{\c{s}}}, \bibinfo{person}{Acyr Locatelli}, {and} \bibinfo{person}{Sara Hooker}.} \bibinfo{year}{2023}\natexlab{}.
\newblock \showarticletitle{Pushing mixture of experts to the limit: Extremely parameter efficient moe for instruction tuning}.
\newblock \bibinfo{journal}{\emph{arXiv preprint arXiv:2309.05444}} (\bibinfo{year}{2023}).
\newblock


\bibitem[Zeng et~al\mbox{.}(2022)]%
        {zeng2022glm}
\bibfield{author}{\bibinfo{person}{Aohan Zeng}, \bibinfo{person}{Xiao Liu}, \bibinfo{person}{Zhengxiao Du}, \bibinfo{person}{Zihan Wang}, \bibinfo{person}{Hanyu Lai}, \bibinfo{person}{Ming Ding}, \bibinfo{person}{Zhuoyi Yang}, \bibinfo{person}{Yifan Xu}, \bibinfo{person}{Wendi Zheng}, \bibinfo{person}{Xiao Xia}, {et~al\mbox{.}}} \bibinfo{year}{2022}\natexlab{}.
\newblock \showarticletitle{GLM-130B: An Open Bilingual Pre-trained Model}. In \bibinfo{booktitle}{\emph{The Eleventh International Conference on Learning Representations}}.
\newblock


\bibitem[Zhang et~al\mbox{.}(2023a)]%
        {zhang2023instruction}
\bibfield{author}{\bibinfo{person}{Shengyu Zhang}, \bibinfo{person}{Linfeng Dong}, \bibinfo{person}{Xiaoya Li}, \bibinfo{person}{Sen Zhang}, \bibinfo{person}{Xiaofei Sun}, \bibinfo{person}{Shuhe Wang}, \bibinfo{person}{Jiwei Li}, \bibinfo{person}{Runyi Hu}, \bibinfo{person}{Tianwei Zhang}, \bibinfo{person}{Fei Wu}, {et~al\mbox{.}}} \bibinfo{year}{2023}\natexlab{a}.
\newblock \showarticletitle{Instruction tuning for large language models: A survey}.
\newblock \bibinfo{journal}{\emph{arXiv preprint arXiv:2308.10792}} (\bibinfo{year}{2023}).
\newblock


\bibitem[Zhang et~al\mbox{.}(2023b)]%
        {yingyingtois}
\bibfield{author}{\bibinfo{person}{Yingying Zhang}, \bibinfo{person}{Xian Wu}, \bibinfo{person}{Quan Fang}, \bibinfo{person}{Shengsheng Qian}, {and} \bibinfo{person}{Changsheng Xu}.} \bibinfo{year}{2023}\natexlab{b}.
\newblock \showarticletitle{Knowledge-Enhanced Attributed Multi-Task Learning for Medicine Recommendation}.
\newblock \bibinfo{journal}{\emph{ACM Trans. Inf. Syst.}}, Article \bibinfo{articleno}{17} (\bibinfo{date}{jan} \bibinfo{year}{2023}), \bibinfo{numpages}{24}~pages.
\newblock


\bibitem[Zhang and Yang(2021)]%
        {zhang2021survey}
\bibfield{author}{\bibinfo{person}{Yu Zhang} {and} \bibinfo{person}{Qiang Yang}.} \bibinfo{year}{2021}\natexlab{}.
\newblock \showarticletitle{A survey on multi-task learning}.
\newblock \bibinfo{journal}{\emph{IEEE Transactions on Knowledge and Data Engineering}} \bibinfo{volume}{34}, \bibinfo{number}{12} (\bibinfo{year}{2021}), \bibinfo{pages}{5586--5609}.
\newblock


\bibitem[Zhao et~al\mbox{.}(2019)]%
        {zhao2019neural}
\bibfield{author}{\bibinfo{person}{Sendong Zhao}, \bibinfo{person}{Ting Liu}, \bibinfo{person}{Sicheng Zhao}, {and} \bibinfo{person}{Fei Wang}.} \bibinfo{year}{2019}\natexlab{}.
\newblock \showarticletitle{A neural multi-task learning framework to jointly model medical named entity recognition and normalization}. In \bibinfo{booktitle}{\emph{Proceedings of the AAAI Conference on Artificial Intelligence}}, Vol.~\bibinfo{volume}{33}. \bibinfo{pages}{817--824}.
\newblock


\bibitem[Zheng et~al\mbox{.}(2024)]%
        {zheng2024harnessing}
\bibfield{author}{\bibinfo{person}{Zhi Zheng}, \bibinfo{person}{Wenshuo Chao}, \bibinfo{person}{Zhaopeng Qiu}, \bibinfo{person}{Hengshu Zhu}, {and} \bibinfo{person}{Hui Xiong}.} \bibinfo{year}{2024}\natexlab{}.
\newblock \showarticletitle{Harnessing Large Language Models for Text-Rich Sequential Recommendation}.
\newblock \bibinfo{journal}{\emph{arXiv preprint arXiv:2403.13325}} (\bibinfo{year}{2024}).
\newblock


\bibitem[Zheng et~al\mbox{.}(2022)]%
        {doctor2022}
\bibfield{author}{\bibinfo{person}{Zhi Zheng}, \bibinfo{person}{Zhaopeng Qiu}, \bibinfo{person}{Hui Xiong}, \bibinfo{person}{Xian Wu}, \bibinfo{person}{Tong Xu}, \bibinfo{person}{Enhong Chen}, {and} \bibinfo{person}{Xiangyu Zhao}.} \bibinfo{year}{2022}\natexlab{}.
\newblock \showarticletitle{DDR: Dialogue Based Doctor Recommendation for Online Medical Service}. In \bibinfo{booktitle}{\emph{Proceedings of the 28th ACM SIGKDD Conference on Knowledge Discovery and Data Mining}} (Washington DC, USA) \emph{(\bibinfo{series}{KDD '22})}. \bibinfo{publisher}{Association for Computing Machinery}, \bibinfo{address}{New York, NY, USA}, \bibinfo{pages}{4592–4600}.
\newblock
\showISBNx{9781450393850}
\urldef\tempurl%
\url{https://doi.org/10.1145/3534678.3539201}
\showDOI{\tempurl}


\bibitem[Zheng et~al\mbox{.}(2023)]%
        {zheng2023interaction}
\bibfield{author}{\bibinfo{person}{Zhi Zheng}, \bibinfo{person}{Chao Wang}, \bibinfo{person}{Tong Xu}, \bibinfo{person}{Dazhong Shen}, \bibinfo{person}{Penggang Qin}, \bibinfo{person}{Xiangyu Zhao}, \bibinfo{person}{Baoxing Huai}, \bibinfo{person}{Xian Wu}, {and} \bibinfo{person}{Enhong Chen}.} \bibinfo{year}{2023}\natexlab{}.
\newblock \showarticletitle{Interaction-aware drug package recommendation via policy gradient}.
\newblock \bibinfo{journal}{\emph{ACM Transactions on Information Systems}} \bibinfo{volume}{41}, \bibinfo{number}{1} (\bibinfo{year}{2023}), \bibinfo{pages}{1--32}.
\newblock


\bibitem[Zhu et~al\mbox{.}(2023)]%
        {zhu2023promptcblue}
\bibfield{author}{\bibinfo{person}{Wei Zhu}, \bibinfo{person}{Xiaoling Wang}, \bibinfo{person}{Huanran Zheng}, \bibinfo{person}{Mosha Chen}, {and} \bibinfo{person}{Buzhou Tang}.} \bibinfo{year}{2023}\natexlab{}.
\newblock \showarticletitle{PromptCBLUE: A Chinese Prompt Tuning Benchmark for the Medical Domain}.
\newblock \bibinfo{journal}{\emph{arXiv preprint arXiv:2310.14151}} (\bibinfo{year}{2023}).
\newblock


\end{thebibliography}

% \clearpage
% \appendix
% \input{7Appendix}

\end{sloppypar}
\end{document}